\documentclass[11pt,final,a4paper]{article}

\usepackage[utf8]{inputenc}
\usepackage{times,latexsym}
\usepackage{bookmark}
\usepackage{natbib}
\usepackage{microtype}

\usepackage[table,dvipsnames]{xcolor}
\usepackage{tcolorbox}
\usepackage{fontawesome5}

\usepackage{subcaption}
\usepackage{tikz}
\usepackage{tikz-3dplot}
\usetikzlibrary{3d, perspective, shapes.geometric, arrows, positioning, calc, decorations.pathreplacing,arrows.meta}

\usepackage[acceptedWithA]{tacl2021v1}

\usepackage{xspace,mfirstuc,tabulary}

\newcommand{\coolblue}[1]{\textcolor{Plum}{#1}}
\newcommand{\warmred}[1]{\textcolor{Red}{#1}}
\newcommand{\green}[1]{\textcolor{OliveGreen}{#1}}
\newcommand{\darkblue}[1]{\textcolor{Blue}{#1}}
\newcommand{\vermillion}[1]{\textcolor{RedViolet}{#1}}
\newcommand{\purplepink}[1]{\textcolor{RubineRed}{#1}}
\newcommand{\yellow}[1]{\textcolor{Sepia}{#1}} 

\newcommand{\guideddecoding}{Top-$k$ Concept-Guided Decoding\xspace}

\newcommand{\methodname}{Frame Representation Hypothesis\xspace}
\newcommand{\methodabbr}{FRH\xspace}
\newcommand{\spacename}{\textit{Semantic Frame Space}\xspace}

\usepackage{amsmath}
\usepackage{amsthm}
\usepackage{amsfonts}
\usepackage{amssymb}
\usepackage{mathtools}
\usepackage{enumitem}
\usepackage{makecell}
\usepackage{mathrsfs}
\usepackage{stmaryrd}
\usepackage{bm}
\usepackage{xparse}

\newcommand*\autoop{\left(}
\newcommand*\autocp{\right)}
\newcommand*\autoob{\left[}
\newcommand*\autocb{\right]}
\AtBeginDocument {%
   \mathcode`( 32768
   \mathcode`) 32768
   \mathcode`[ 32768
   \mathcode`] 32768
   \begingroup
       \lccode`\~`(
       \lowercase{%
   \endgroup
       \let~\autoop
   }\begingroup
       \lccode`\~`)
       \lowercase{%
   \endgroup
       \let~\autocp
   }\begingroup
       \lccode`\~`[
       \lowercase{%
   \endgroup
       \let~\autoob
   }\begingroup
       \lccode`\~`]
       \lowercase{%
   \endgroup
       \let~\autocb
}}
\makeatletter
\AtBeginDocument {%
          \def\resetMathstrut@{%
           \setbox\z@\hbox{\the\textfont\symoperators\char40}%
           \ht\Mathstrutbox@\ht\z@ \dp\Mathstrutbox@\dp\z@}%
}%
\makeatother

\DeclareMathOperator{\logit}{logit}
\DeclareMathOperator{\Cov}{Cov}
\DeclareMathOperator{\tr}{tr}

\DeclareMathOperator{\ray}{\mathcal{R}}
\DeclareMathOperator{\diffframe}{\mathbb{D}}

\newcommand{\mat}[1]{\mathbf{#1}}
\newcommand{\subspacemat}[2][]{\left[\mathbf{#2}_{#1}\right]}
\newcommand{\subspace}[1]{\mathcal{#1}}

\newcommand{\reals}{\mathbb{R}}
\newcommand{\numbers}{\mathsf{k}}
\newcommand{\dimension}{\mathsf{d}}
\DeclareMathOperator{\rank}{rank}

\newcommand{\vectorspace}{\reals^\dimension}
\newcommand{\matrixspace}{\reals^{\dimension \times \numbers}}

\newcommand{\flagtypesize}{\mathsf{q}}

\DeclareMathOperator{\Stiefel}{\mathrm{St}}

\newcommand{\stiefelmanifold}[1]{\Stiefel(\numbers_{#1}, \dimension)}

\newcommand{\completestiefelmanifold}[1]{\mathbf{CSt}({#1}, \dimension)}

\DeclarePairedDelimiterX\innerproduct[2]{\langle}{\rangle}{#1 \delimsize, #2}
\DeclareMathOperator{\innerproductmatrix}{\mat{M}}

\newcommand{\vect}[1]{\mathbf{#1}}

\DeclareMathOperator{\Frobenius}{\mathrm{F}}
\DeclareMathOperator{\Projection}{\mathsf{P}}
\DeclareMathOperator{\Procrustes}{\mathbf{P}}

\DeclareMathOperator{\dist}{\mathrm{d}}
\DeclareMathOperator{\distcf}{\mathrm{d}_{\Procrustes}}
\DeclareMathOperator{\distcfasym}{\mathrm{d}^\ast_{\Procrustes}}

\DeclareMathOperator{\corr}{\rho}

\newcommand{\set}[1]{\{{#1}\}}
\newcommand{\setsize}{\mathrm{n}}
\newcommand{\setwithcount}[3][\setsize]{\set{{#2}_{#3}}_{#3=1}^{#1}}
\newcommand{\setsum}[1]{\sum_{#1=1}^{\setsize}}

\DeclareMathOperator{\expectation}{\mathbb{E}}

\newcommand{\token}[1]{\mathtt{#1}}
\newcommand{\sentencelength}{\mathsf{t}}

\NewDocumentCommand{\tokensequence}{ O{\sentencelength} m }{
  \token{#2}_\mathnormal{1}, \token{#2}_\mathnormal{2}, \dots, \token{#2}_\mathnormal{#1}
}

\newcommand{\vocabulary}{\mathcal{V}}
\newcommand{\vocabularysize}{\abs{\vocabulary}}

\newcommand{\embeddingspace}{\mathcal{E}}
\newcommand{\embedding}{\vect{e}}
\newcommand{\embeddingmat}{\mat{W}_\embeddingspace}
\newcommand{\embeddingmatrixspace}{\reals^{\dimension \times \vocabularysize}}

\newcommand{\unembeddingspace}{\mathcal{U}}
\newcommand{\unembedding}{\vect{u}}
\newcommand{\unembeddingmat}{\mat{W}_\unembeddingspace}
\newcommand{\unembeddingmatrixspace}{\reals^{\vocabularysize \times \dimension}}

\newcommand{\inputfeaturespace}{\mathcal{H}}
\newcommand{\feature}{\vect{h}}
\newcommand{\featuremat}{\mat{H}}

\newcommand{\llmspaces}{$\inputfeaturespace$ and $\unembeddingspace$ }

\newcommand{\padtoken}{\token{[PAD]}}
\newcommand{\eostoken}{\token{[EOS]}}

\newcommand{\wordmat}[1]{\mat{W}_\mathnormal{#1}}

\newcommand{\word}[1]{\textsf{#1}}

\newcommand{\synset}[1]{\texttt{#1}}
\newcommand{\concept}[1]{\synset{#1}}
\newcommand{\conceptcount}{\mathsf{s}}
\newcommand{\conceptcoef}[1]{a_{#1}}
\newcommand{\conceptvect}[1]{\vect{s}_{\mathnormal{#1}}}

\newcommand{\conceptmat}[1]{\mat{S}_{\mathnormal{#1}}}

\newcommand{\makematrix}[1]{\begin{pmatrix} #1 \end{pmatrix}}
\newcommand{\conceptmatdef}{\mat{S} = \makematrix{\conceptvect{1} & \conceptvect{2} & \dots & \conceptvect{\numbers}}}
\newcommand{\wordmatdef}[1]{\mat{W}_{#1} = \makematrix{\unembedding(\token{w}_{{#1}1}) & \unembedding(\token{w}_{{#1}2}) & \dots & \unembedding(\token{w}_{{#1}\numbers_{{#1}}})}}

\newcommand{\onehalf}{\text{\textonehalf}}

\newcommand{\appropto}{\mathrel{\vcenter{
  \offinterlineskip\halign{\hfil$##$\cr
    \propto\cr\noalign{\kern2pt}\sim\cr\noalign{\kern-2pt}}}}}

\DeclareMathOperator*{\argmax}{arg\,max}
\DeclareMathOperator*{\argmin}{arg\,min}

\DeclarePairedDelimiter\abs{\lvert}{\rvert}
\DeclarePairedDelimiter\norm{\lVert}{\rVert}

\makeatletter
\let\oldabs\abs
\def\abs{\@ifstar{\oldabs}{\oldabs*}}
\let\oldnorm\norm
\def\norm{\@ifstar{\oldnorm}{\oldnorm*}}
\makeatother

\theoremstyle{remark}

\theoremstyle{remark}

\theoremstyle{definition}

\theoremstyle{definition}

\theoremstyle{definition}
\newtheorem{postulate}{Postulate}[section]

\theoremstyle{definition}

\newtheorem{lemma}{Lemma}[section] 
\newtheorem{proposition}{Proposition}[section]

\usepackage{xspace}

\makeatletter
\DeclareRobustCommand\onedot{\futurelet\@let@token\@onedot}
\def\@onedot{\ifx\@let@token.\else.\null\fi\xspace}

\def\eg{\emph{e.g}\onedot} 
\def\ie{\emph{i.e}\onedot} 
\def\cf{\emph{c.f}\onedot}

\makeatother
\definecolor{usercolor}{RGB}{230,230,230}
\definecolor{assistantcolor}{RGB}{240,240,255}
\definecolor{parametercolor}{RGB}{255,240,240}

\newcommand{\highlight}[2][usercolor]{%
  \colorbox{#1}{#2}%
}

\newcommand{\user}[1]{
  \begin{tcolorbox}[
    colback=usercolor,
    colframe=gray!50,
    arc=2mm,
    boxrule=0.5pt,
    left=1pt,
    right=1pt,
    top=1pt,
    bottom=1pt,
    width=\linewidth-4em,
  ]
    \begin{minipage}[c][3em][c]{2em}
      \centering
      \faUser
    \end{minipage}%
    \hfill
    \begin{minipage}[c]{\dimexpr\linewidth-2.5em\relax}
      #1
    \end{minipage}
  \end{tcolorbox}
}

\newcommand{\assistant}[3]{
  \begin{tcolorbox}[
    colback=assistantcolor,
    colframe=gray!50,
    arc=0mm,
    boxrule=0.5pt,
    left=2pt,
    right=2pt,
    top=2pt,
    bottom=2pt
  ]
    \begin{minipage}[c]{\dimexpr\linewidth-7.5em\relax}
      #1
    \end{minipage}%
    \hfill
    \begin{minipage}[c][4em][c]{7em}
      \centering
      \faRobot\\
      \begin{tcolorbox}[
        colback=parametercolor,
        colframe=gray!50,
        arc=0mm,
        boxrule=0.5pt,
        left=1pt,
        right=1pt,
        top=1pt,
        bottom=1pt,
        width=7em
      ]
        \centering
        \tiny #2
      \end{tcolorbox}
    \end{minipage}
  \end{tcolorbox}
}
\usepackage{graphicx}
\usepackage{arydshln}
\usepackage{colortbl}
\usepackage{tikz} 
\usepackage{adjustbox} 

\graphicspath{{./images/}}

\setkeys{Gin}{draft=false}
\usepackage{cleveref}

\hypersetup{
    draft=false,
    colorlinks=true,
    linkcolor=blue,
    filecolor=magenta,      
    urlcolor=cyan,
    pdfpagemode=FullScreen,
    linkbordercolor={0 0 0},
    pdfborderstyle={/S/U/W 0.0},
    pdftitle={Flag Representation Hypothesis},
    pdfauthor={PHVV},
}
\crefname{postulate}{postulate}{postulates}
\Crefname{postulate}{Postulate}{Postulates}

\crefname{definition}{definition}{definitions}
\Crefname{definition}{Definition}{Definitions}

\crefname{lemma}{lemma}{lemmas}
\Crefname{lemma}{Lemma}{Lemmas}

\crefname{proposition}{proposition}{propositions}
\Crefname{proposition}{Proposition}{Propositions}

\crefname{theorem}{theorem}{theorems}
\Crefname{theorem}{Theorem}{Theorems}

\crefname{corollary}{corollary}{corollaries}
\Crefname{corollary}{Corollary}{Corollaries}

\newif\iftaclinstructions
\taclinstructionsfalse 
\iftaclinstructions

\renewcommand{\confidential}{}
\renewcommand{\anonsubtext}{(No author info supplied here, for consistency with
TACL-submission anonymization requirements)}
\newcommand{\instr}
\fi

\iftaclpubformat 

\else

\fi

\title{\methodname: Multi-Token LLM Interpretability and Concept-Guided Text Generation}

\author{
  Pedro H. V. Valois$^\ast$,
  Lincon S. Souza$^\dagger$,
  Erica K. Shimomoto$^\dagger$,
  Kazuhiro Fukui$^\ast$
  \\
  $^\ast$University of Tsukuba,
  $^\dagger$National Institute of Advanced Industrial Science and Technology (AIST)
  \\
}

\date{}

\begin{document}


\maketitle

\begin{abstract}
    Interpretability is a key challenge in fostering trust for Large Language Models (LLMs), which stems from the complexity of extracting reasoning from model's parameters. We present the \methodname, a theoretically robust framework grounded in the Linear Representation Hypothesis (LRH) to interpret and control LLMs by modeling multi-token words. Prior research explored LRH to connect LLM representations with linguistic concepts, but was limited to single token analysis. As most words are composed of several tokens, we extend LRH to multi-token words, thereby enabling usage on any textual data with thousands of concepts. To this end, we propose words can be interpreted as frames, ordered sequences of vectors that better capture token-word relationships. Then, concepts can be represented as the average of word frames sharing a common concept. We showcase these tools through \guideddecoding{}, which can intuitively steer text generation using concepts of choice. We verify said ideas on Llama 3.1, Gemma 2, and Phi 3 families, demonstrating gender and language biases, exposing harmful content, but also potential to remediate them, leading to safer and more transparent LLMs. Code is available at \url{https://github.com/phvv-me/frame-representation-hypothesis.git}
\end{abstract}
\section{Introduction}

Interpretability in deep learning aims to elucidate how neural networks derive predictions. As models grow complex, understanding internal mechanisms gets challenging. By identifying factors contributing to the output, we can foster trust, safety, fairness and improve capabilities~\cite{hooker_benchmark_2019}.

The goal of this study is to enhance the interpretability and control of LLMs via the encoding of human-comprehensible \textit{linguistic concepts}. LLMs represent text through tokens, which can be a word, part of a word, or even a character, as per models' design. In contrast, humans better understand text through \textit{concepts}, cognitive symbols that depict reality, often grouping related objects, events, or further abstractions based on shared characteristics. Our purpose is to provide tools to represent concepts within LLMs, allowing output explanations that are suited for our mental models.

Prominent works that offer such tools are based on the Linear Representation Hypothesis (LRH): it suggests linear operations on token vectors can explain model behavior, with concepts represented as vectors in LLM feature space~\cite{templeton2024scaling}. For instance, we can identify the concept \synset{\yellow{female}} as the average of token vectors like $f(\word{\yellow{woman}})$ or $f(\word{\yellow{queen}})$. Thus, token vectors encode more than just lexical data: they also represent linguistic concepts~\cite{Mikolov2013LinguisticRI}.

\begin{figure}[tb]
\centering
\includegraphics[width=0.6\linewidth]{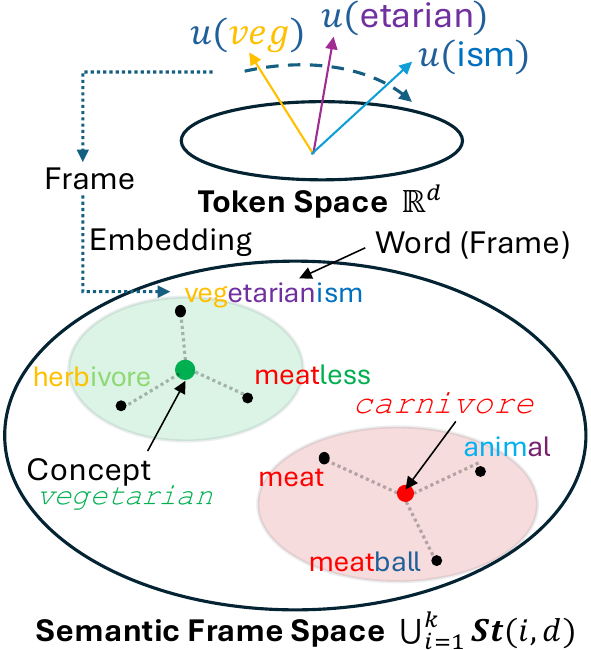}
\caption{\methodname Overview: Tokens are vectors, which combine into words as multi-dimensional frames. In turn, Concept Frames are centroids of word sets.}
\label{fig:general_diagram}
\end{figure}

Nevertheless, LRH's concepts are 1-dimensional, constraining them to single-token words, which are a minor fraction of any given language~\cite{Bau2020UnderstandingTR}. As exemplified in \Cref{fig:general_diagram}, a concept like \synset{\green{vegetarian}} is exclusively linked to multi-token words, such as \word{\vermillion{meat}\green{less}} or \word{\warmred{herb}\green{ivore}}, meaning multiple vectors are required to represent it. Since most words are constituted of several tokens, 1-dimensional structures prevent LRH application in most interpretability tasks effectively.

To address this shortcoming of LRH, we propose a new framework for LLM interpretability, based on a key empirical observation about the nature of LLMs: our experiments show that over 99\% of words among several languages are composed of linearly independent token vectors. This allows us to address multi-token words by proposing the \methodname (\methodabbr), which assumes words are ordered sequences of independent vectors -- mathematically identified as frames.

Starting from this postulate, we develop a mathematical framework to represent words and concepts as frames; we define a \spacename, and equip it with a correlation between frames that preserves the token whitening mechanism introduced by LRH~\cite{park2023linear}, allowing us to measure frames semantic relationship. Following \Cref{fig:general_diagram}, the frame representation lets us identify words, such as \word{\vermillion{meat}\green{less}}, \word{\vermillion{meat}} and \word{\vermillion{meat}\darkblue{ball}} as distinct geometrical objects, although they have tokens in common. Then, we can compute Concept Frames like \synset{\green{vegetarian}} or \synset{\vermillion{carnivore}} as the centroid for a set of words sharing that concept.

Furthermore, we introduce \guideddecoding{} (\Cref{fig:top_k_guided_decoding}), which controls text generation by selecting the tokens which maximize a chosen concept. For example, the concept \synset{\green{vegetarian}} would guide input \word{\darkblue{I like}} to \word{\darkblue{I like} \green{fruits}} if the top-$3$ options were \word{\yellow{beef}}, \word{\coolblue{football}} or \word{\green{fruits}}. This algorithm aligns model outputs with desired concepts, a practical prototype for \methodabbr that allows meaningful LLM understanding.

To that end, we leverage the Open Multilingual WordNet (OMW)~\cite{bond2013linking} as a source of synonyms to build concepts. We use over 50M words among multiple languages to build over 100,000 Concept Frames, enabling rich model understanding in a diverse yet inexpensive manner. 

In short, \methodabbr formally extends the LRH to multi-token words. We show its validity both from the theoretical and empirical points of view.

Our primary contributions are as follows:

\begin{enumerate}
    \item \methodname as an extension of LRH to multi-token words by defining them as Frames, thereby addressing the limitations of single-token representations.
    \item Proposal of Concept Frames to represent linguistic concepts from a set of Word Frames.
    \item Development of \guideddecoding{}, a proof-of-concept application to steer text generation using chosen concepts and expose model biases or potential vulnerabilities.
\end{enumerate}

\section{Related Work}

We briefly review LLM interpretability, controllable text generation and Frame usage in the field.

\paragraph{Language Models Interpretability}

The widespread adoption of LLMs brought attention to the need of understanding their inner-workings, risks, and limitations. Several studies identified a common property to these models that became known as the Linear Representation Hypothesis, encoding model knowledge as vectors~\cite{Mikolov2013LinguisticRI}, and enabling model explanation and editing~\cite{Wang2023KnowledgeEF}. Also, the Superposition Hypothesis (SH) assumes specialized information is superimposed in LLM feature spaces. These ideas underpin Sparse Autoencoders, which learn dictionaries of interpretable concepts to decipher model behavior~\cite{elhage2022superposition}, whereas our proposal uses WordNet~\cite{miller1995wordnet} to map learned representations to concepts. For a comprehensive survey, see \citet{Ferrando2024APO}.

\paragraph{Controllable Text Generation}

LLMs can use various decoding strategies for inference. Beam search~\cite{Jurafsky2000SpeechAL, Graves2012SequenceTW} can improve quality but risk cycles. Top-$k$~\cite{Fan2018HierarchicalNS} and nucleus sampling~\cite{Holtzman2019TheCC} introduce randomness for diversity. Other techniques offer specific controls, such as poetry generation~\cite{ghazvininejad-etal-2017-hafez}, attribute maximization~\cite{Krause2020GeDiGD}, style optimization~\cite{Khalifa2020ADA}, reasoning paths~\cite{Wang2022SelfConsistencyIC}, diverse decisions~\cite{Yao2023TreeOT}, and self-evaluation~\cite{Kadavath2022LanguageM, Xie2023SelfEvaluationGB}. LRH enables steering through linear interventions for knowledge edition~\cite{belrose2024leace,singh2024mimic} or harmfulness~\cite{bai2022training}, while our proposal guides text generation by maximizing a certain concept.

\paragraph{Subspaces and Frames in Machine Learning}

Subspaces have been extensively used for dimensionality reduction, feature extraction~\cite{fukui2015differencesubspace,9760096}, classification~\cite{watanabe1967evaluation}, image interpretability~\cite{Valois2023OcclusionSA}, and modelling text sentences~\cite{cancedda-2024-spectral,Shimomoto2021TextCB}. Their invariance to selection of basis is advantageous for representing clusters, but limiting for ordered structures. Frames are sequences of vectors that model well redundant and oriented data, being applied in error correction~\cite{frames01}, signal decomposition~\cite{casazza2013introduction} and optimization problems~\cite{mankovich2023chordal,Chaudhry2020ContinualLI}. To the best of our knowledge, frames have never been applied in NLP to model words and concepts as proposed here.

\section{Preliminary} \label{sec:preliminary}

\begin{figure}[tb]
    \centering
    \includegraphics[width=\linewidth]{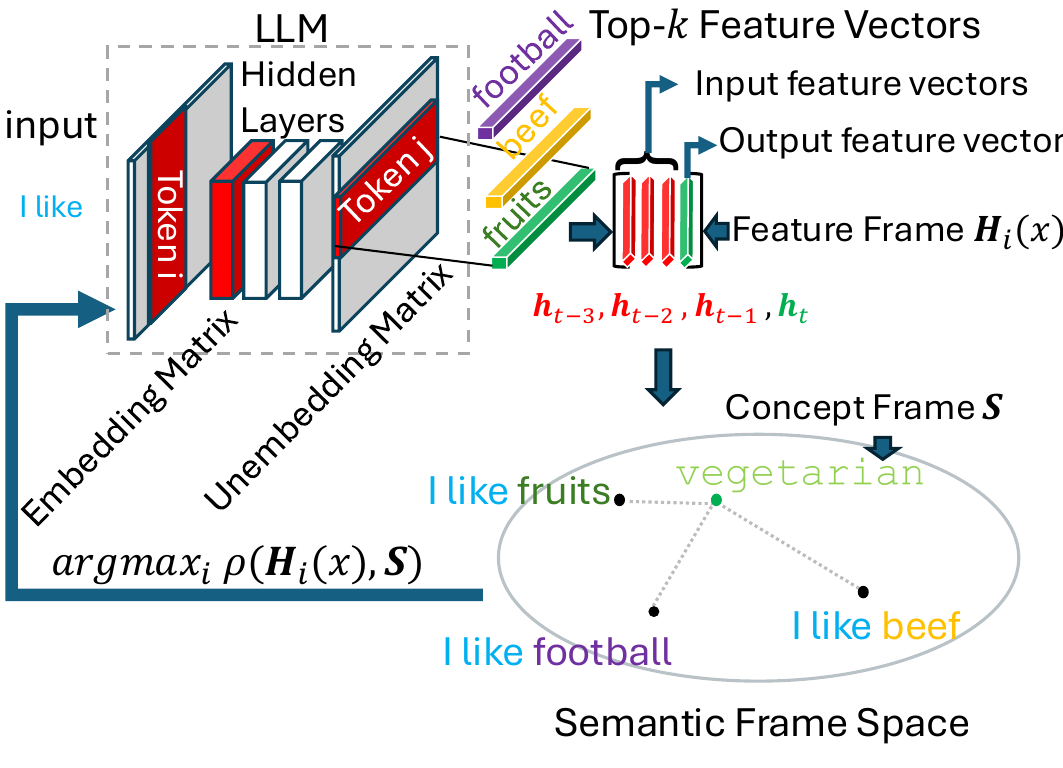}
    \caption{\guideddecoding Overview: Top-$k$ sentence candidates are derived from the model logits, and we chose the one which maximizes the correlation with the target Concept Frames. The process is repeated in a loop until the desired number of tokens is reached.}
    \label{fig:top_k_guided_decoding}
\end{figure}

In this section, we introduce the necessary background to our proposal. Throughout this work, we denote vectors as bold lowercase letters, \eg, $\vect{v}$; matrices as bold uppercase letters, \eg, $\mat{M}$; monospace lowercase letters for tokens, \eg, $\token{x}$; spaces with calligraphic letters, \eg, $\unembeddingspace$; words with sans serif uppercase uppercase letters, \eg, \word{W}, and concepts with monospace uppercase letters, \eg, \concept{C}. 

\subsection{Frames}

A $\numbers$-frame is a sequence of $\numbers$ linearly independent vectors in $\vectorspace$, represented by $\mat{F} \in \matrixspace, \rank(\mat{F}) = \numbers$. The set of all $\numbers$-frames in $\vectorspace$ constitutes the non-compact Stiefel manifold~$\stiefelmanifold{}$. Manifolds are structures in which distance, geodesics and more may be defined, so we can compute geometrically meaningful relationships between distinct frames~\cite{edelman1998geometry}. 
The set of all frames up to rank $\flagtypesize$ forms the $\flagtypesize$-complete Stiefel manifold $\completestiefelmanifold{\flagtypesize} = \bigcup_{i = 0}^{\flagtypesize} \Stiefel(i, \dimension)$ -- a disjoint union of Stiefel manifolds -- where the null frame $\Stiefel(0, \dimension) \equiv \varnothing$ is defined as our space's origin. Intuitively, $\completestiefelmanifold{\flagtypesize}$ is a stratified structure, so its base is the null frame.




\subsection{Rays}

A ray $\ray$ is a directed half-infinite line, also known as 1-dimensional convex cone, half-line, or axis~\cite{boyd2004convex}, defined by a vector $\vect{v} \in \vectorspace$ and its scalar multiples $\ray(\vect{v}) = \{ \alpha \vect{v} \mid \alpha \ge 0 \}$. A ray is represented by a normalized vector $\vect{v}' = \vect{v}/\norm{\vect{v}}$, which is also a point in $\Stiefel(1, \dimension)$. Rays differ from subspaces by their orientation, so a single dimensional subspace contains two rays, and their correlation is measured as the cosine of the angle $\theta$ between their normalized vectors~\cite{angle-math}

\begin{equation}\label{eq:ray-corr}
    \corr(\ray(\vect{v}),\ray(\vect{u})) = \frac{\innerproduct{\vect{v}}{\vect{u}}}{\norm{\vect{v}}\norm{\vect{u}}} = \vect{v}'^\top \vect{u}' = \cos\theta .
\end{equation}

Moreover, notice the correlation shown in \Cref{eq:ray-corr} is connected to the inner product -- projection of one vector onto another. In this work, we use the term ``projection'' when calculating correlation with unnormalized vectors.

\subsection{Large Language Models} \label{sec:llm}

LLM models process text by converting it into a sequence of tokens, \textit{embedding} them into its own vector space and processing this sequence of vectors through its hidden layers to a final vector representation, which is \textit{unembedded} into the most likely token to continue the input sentence. A simple version of such pipeline is illustrated in \Cref{fig:top_k_guided_decoding}.

A token is a single element of a textual sequence, represented by a number $\token{x} \in \vocabulary$ in a predefined vocabulary $\vocabulary \subset \mathbb{Z}^+$. In that sense, the model's tokenizer converts text input $x$ into token $\sentencelength$-tuple $(\tokensequence{x}) \in \vocabulary^\sentencelength$. The LLM then starts in the \textit{embedding} layer,
which maps each token number $\token{a} \in \vocabulary$ to an unique \textit{embedding} vector $\embedding(\token{a}) \in \embeddingspace \cong \vectorspace$, each of which is a column of the \textit{embedding} matrix $\embeddingmat \in \embeddingmatrixspace$. Therefore, the output of this layer is the $\sentencelength$-tuple of \textit{embedding} vectors $\embedding(x) = (\embedding(\token{x}_1), \embedding(\token{x}_2), \dots, \embedding(\token{x}_\sentencelength))$. Next, $\embedding(x)$ is processed by the DNN hidden transformer layers into the feature vector $\feature(x) = \feature(\embedding(\token{x}_1), \embedding(\token{x}_2), \dots, \embedding(\token{x}_\sentencelength)) \in \inputfeaturespace \cong \vectorspace$. 

Then, the LLM converts $\feature(x)$ into a token number. The \textit{unembedding} vector of token $\token{b} \in \vocabulary$ is $\unembedding(\token{b}) \in \unembeddingspace \cong \vectorspace$, a row of the \textit{unembedding} matrix $\unembeddingmat \in \unembeddingmatrixspace$, which also identifies each token to a unique vector in high-dimensional space $\unembeddingspace$. 

Finally, the probability of a token $\token{y} \in \vocabulary$ being next in a text sentence $x$ is determined with softmax

\begin{equation} \label{eq:llm-prob}
    p(\token{y}|x) \propto \exp(\unembedding(\token{y})^\top \feature(x)).
\end{equation}

In practice, the space dimension $\dimension$ can range from 1024 to 16384, while the vocabulary $\vocabulary$ usually contains from 50,000 to 300,000 tokens.


\subsection{Linear Representation Hypothesis} \label{sec:lrh}

We now concentrate the discussion into the geometry of \llmspaces and their relationships. With that in mind, \citet{park2023linear} defined the \textit{ray representation} of a concept $\concept{C}$ as the ray $\ray(\conceptvect{\concept{C}}') \subset \unembeddingspace$ of vector $\conceptvect{\concept{C}}' \in \unembeddingspace$. The correlation $\corr$ of concepts $\concept{A}, \concept{B}$ serves as a linear probe for model understanding
\begin{equation}\label{eq:lrh-inner-product}
    \corr(\concept{A}, \concept{B}) = \innerproduct{\conceptvect{\concept{A}}'}{\conceptvect{\concept{B}}'} = \conceptvect{\concept{A}}'^\top \mat{M} \conceptvect{\concept{B}}',
\end{equation}
where $\mat{M} = \Cov^{-1}(\unembeddingmat)$ is a whitening matrix that defines the LRH inner product, placing unrelated concepts as orthogonal to each other.


Hereafter, concepts connect through linear operations and are computed as the normalized mean of counterfactual pairs difference vectors
\begin{align} 
    \label{eq:un-repr-1}
    \unembedding_{\concept{C}}' &= \sum_{i}^{n_{\concept{C}}} (\unembedding_i({\concept{C}}=1) - \unembedding_i({\concept{C}}=0)), \\
    \label{eq:un-repr-2}
    \conceptvect{{\concept{C}}}' &= \frac{\unembedding_{\concept{C}}'}{\norm{\unembedding_{\concept{C}}'}},
\end{align}
where $\unembedding_i({\concept{C}}=1), \unembedding_i({\concept{C}}=0)$ is a counterfactual token pair, so ${\concept{C}} = 1$ indicates one concept direction while ${\concept{C}} = 0$ its opposite, \eg, concept ${\text{\concept{\purplepink{English}}}\Rightarrow\text{\concept{\darkblue{Spanish}}}} ({\concept{C}} = 0)$ is computed using difference vectors like $\unembedding(\text{\word{\purplepink{good}}}) - \unembedding(\text{\word{\darkblue{bueno}}})$, $\unembedding(\text{\word{\purplepink{bad}}}) - \unembedding(\text{\word{\darkblue{malo}}})$, while ${\text{\concept{\darkblue{Spanish}}}}\Rightarrow\text{\concept{\purplepink{English}}} ({\concept{C}} = 1)$ is the opposite vector.

\subsection{WordNet}~\label{sec:wordnet}

The Open Multilingual WordNet (OMW) is a collaborative project that intersects cognitive psychology, linguistics and computer science to create an interconnected network of lexical databases~\cite{bond2013linking, fellbaum1998wordnet, harabagiu1999wordnet}. At its core are \textbf{synsets} and \textbf{lemmas}. A synset, short for ``synonym set'', is a group of words or phrases that may share the same meaning. For example, \{\word{\coolblue{car}}, \word{\coolblue{automobile}}, \word{\coolblue{auto}}\} forms a synset, which can be uniquely identified as \synset{\coolblue{car.n.01}}, \ie, the 1st dictionary meaning of the word \word{\coolblue{car}} as a noun~\cite{miller1990introduction}. 

A lemma, on the other hand, is a canonical form, \eg, \word{\warmred{run}}, \word{\warmred{runs}}, \word{\warmred{ran}}, and \word{\warmred{running}} are all represented by the lemma \word{\warmred{run}}~\cite{fellbaum2010wordnet}. In OMW, synsets from different languages are linked to their equivalent English ones, allowing for cross-lingual connections. Therefore, the English synset \synset{\coolblue{car.n.01}} will also include Spanish lemmas, such as \word{\coolblue{coche}} or \word{\coolblue{automóvil}}~\cite{bond2012survey}. Each lemma can belong to multiple synsets, reflecting its different meanings, making OMW a powerful NLP tool~\cite{Wagner2010StevenBE}.
\newcommand{\tokenad}{$\word{\vermillion{ad}}$}
\newcommand{\tokenmit}{$\word{\darkblue{mit}}$}
\newcommand{\wordadmit}{\tokenad\tokenmit}
\newcommand{\wordmitad}{\tokenmit\tokenad}

\newcommand{\synsetad}{\synset{\vermillion{ad.n.01}}}
\newcommand{\synsetmit}{\synset{\darkblue{myth.n.01}}}
\newcommand{\synsetadmit}{\synset{\green{admit.v.01}}}
\newcommand{\synsetmitad}{\synset{\purplepink{half.n.02}}}

\newcommand{\tokenone}{$\word{\green{1}}$}
\newcommand{\tokennine}{$\word{\coolblue{9}}$}
\newcommand{\tokeneight}{$\word{\warmred{8}}$}
\newcommand{\tokenzero}{$\word{\darkblue{0}}$}
\newcommand{\tokenS}{$\word{\purplepink{s}}$}

\newcommand{\wordnineteeneighties}{\tokenone\tokennine\tokeneight\tokenzero\tokenS}
\newcommand{\wordeightynineties}{\tokenone\tokeneight\tokennine\tokenzero\tokenS}

\section{\methodname} \label{sec:methods}

In this section, we introduce our theoretical framework. Proofs are provided in \Cref{sec:proofs}.

\subsection{Linear Decomposition of Tokens} \label{sec:linear-decomposition}

\begin{figure}[tb]
    \centering
    \includegraphics[width=0.95\linewidth]{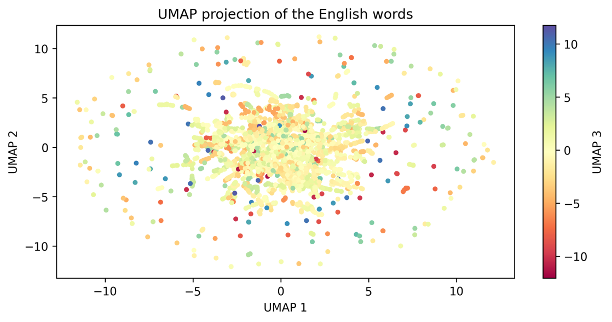}
    \caption{Uniform Manifold Approximation and Projection (UMAP)~\cite{McInnes2018UMAPUM} of the 10k most frequent single-token English words for Gemma 2. While some points are clearly separated, others  overlap due to the Superposition Hypothesis (SH). For example, \tokenad{} is a token in the unrelated words \word{\vermillion{advertisement}}, \word{\darkblue{admit}}, \word{\purplepink{adventure}}, etc., while \word{\yellow{restaurant}} is a single token and it is not found in other words.}
    \label{fig:umap-english}
\end{figure}

LRH posits concepts are linearly encoded within LLMs feature spaces. Moreover, the Superposition Hypothesis (SH) suggests models encode information in a \textit{superposition of concepts} because the number of possible concepts significantly exceeds the space dimensionality~\cite{elhage2022superposition}, a phenomenon visible in \Cref{fig:umap-english}. This is mathematically expressed as a linear combination of vectors, formalized at \Cref{ax:lin-comb-concept}.

\begin{postulate} \label{ax:lin-comb-concept}
    Let $\unembedding(\token{y})$ be the unembedding representation of token $\token{y} \in \vocabulary$, then it is a linear combination of \textit{concept vectors} $\conceptvect{}$
    \begin{equation} \label{eq:lin-comb-concept}
        \unembedding(\token{y}) - \unembedding_\token{0} = \sum_{i}^{\conceptcount} a_i \conceptvect{i},
    \end{equation}
    where $a_i \in \reals$, $\conceptcount$ is the number of all concepts known by the model, and $\unembedding_\token{0}$ is a \textit{meaningless} vector -- an offset element from the fact not all tokens might hold meaning, \eg, $\padtoken$ or $\eostoken$, implying we need to remove the \textit{meaningless} part of each token vector. Heuristically, $\unembedding_\token{0}$ should be the unembedding vector average, so that $\expectation[\unembedding(\token{y}) - \unembedding_\token{0}] = 0$. Also, $\setwithcount{\conceptvect{}}{i}$ is not a basis: words can be grouped in several ways, \eg, antonyms or synonyms, making concepts interdependent.
\end{postulate}

Therefore, we can extract a concept of choice by averaging tokens sharing that concept. Let $\setwithcount{\token{y}}{j}$ be a set of tokens sharing a common concept $\conceptvect{}$, we estimate the concept as the token average\footnote{hereafter consider all tokens to be already debiased}

\begin{equation} \label{lemma:concept-estimation}
    \conceptvect{} \appropto \setsum{j} \unembedding(\token{y}_j) - \unembedding_0,
\end{equation}



\subsubsection{Combined Concepts} \label{sec:combined-concepts}


We connect Concept Estimation (\ref{lemma:concept-estimation}) to the ray $\ray(\conceptvect{{\concept{C}}}')$ of a concept ${\concept{C}}$ by separating \Cref{eq:un-repr-1} into two sums, each its own concept. Therefore, a concept ${\concept{C}}$ has representation $\ray(\conceptvect{{\concept{C}}}')$, where $\conceptvect{{\concept{C}}}'$ is a normalized counterfactual concept pair difference, 
\begin{equation} \label{prop:concept-der}
    \conceptvect{{\concept{C}}}' = \frac{\conceptvect{{\concept{C}}=1} - \conceptvect{{\concept{C}}=0}}{\norm{\conceptvect{{\concept{C}}=1} - \conceptvect{{\concept{C}}=0}}},
\end{equation}
indicating some concepts are formed by other concepts. For example, a set of tokens sharing the meaning of \concept{\yellow{female}} builds $\conceptvect{\text{\concept{\yellow{female}}}}$, while another sharing the concept of \concept{\green{male}} builds $\conceptvect{\text{\concept{\green{male}}}}$, forming
\begin{equation}\label{eq:combined-concept-difference} \conceptvect{\text{\concept{\green{male}}}\Rightarrow\text{\concept{\yellow{female}}}}' \propto \conceptvect{\text{\concept{\yellow{female}}}} - \conceptvect{\text{\concept{\green{male}}}},
\end{equation}
which leads us to understand some concepts as building blocks for Combined Concepts.

\subsection{Generalizing from Tokens to Words}~\label{sec:gen-frh}

\begin{figure}[tb]
    \centering
    \includegraphics[width=\linewidth]{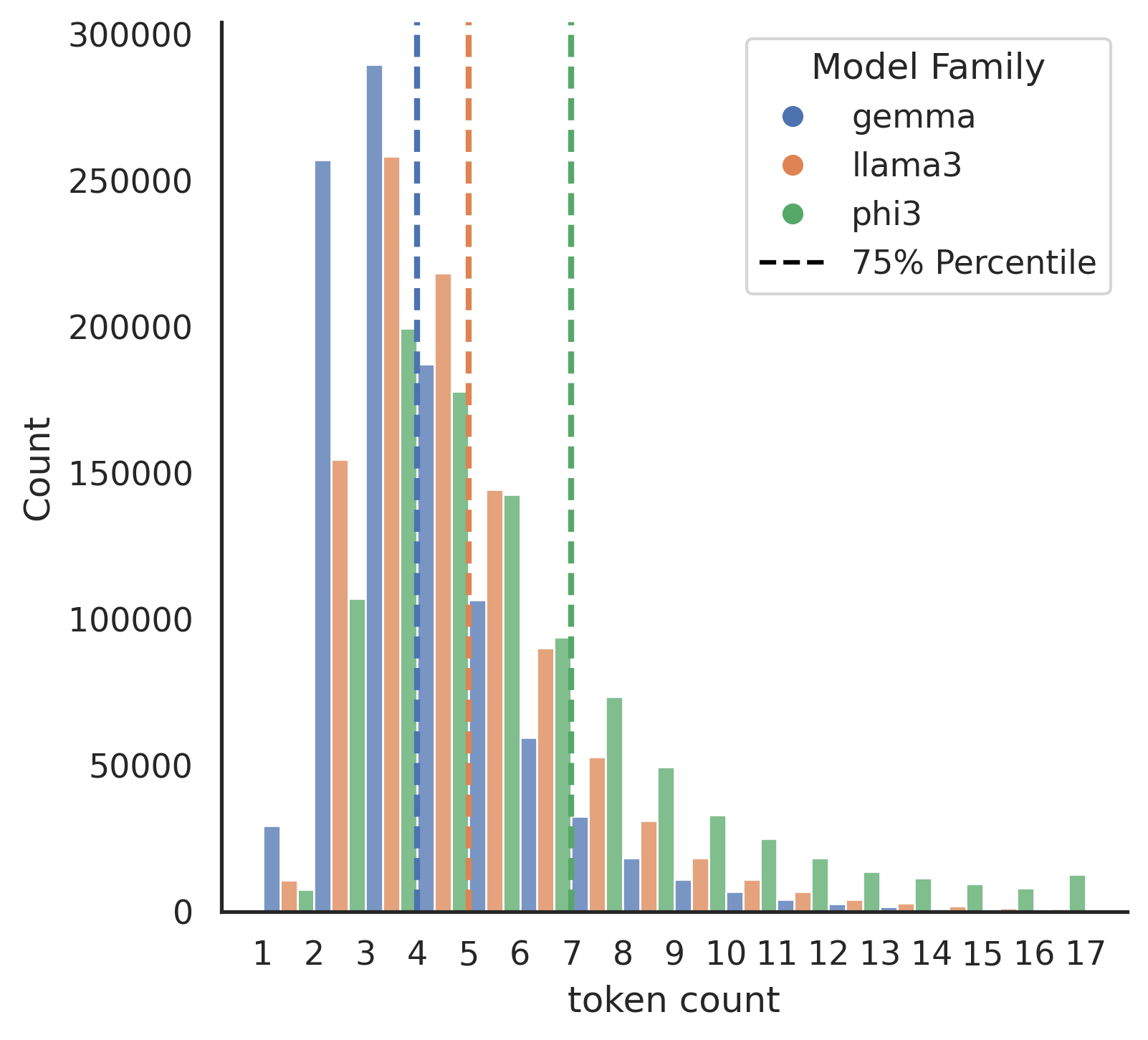}
    \caption{Histogram of lemma token count among all OMW lemmas. The dashed vertical bar indicates the 75\% percentile for each model family.}
    \label{fig:hist-token-count}
\end{figure}

The previous discussion can leverage WordNet to determine concepts. WordNet's structure overlaps with LLM representations~\cite{Moskvoretskii2024TaxoLLaMAWM,Moskvoretskii2024AreLL,park2024geometry}, and OMW synsets are sets of multilingual lemmas sharing a meaning, making it well suited for \Cref{lemma:concept-estimation}. Nevertheless, \Cref{sec:linear-decomposition} only deals with single-token words, which accounts for less than 1\% of all OMW lemmas on most LLMs, significantly limiting estimated concepts quality (\cf \Cref{fig:hist-token-count}).

In order to make LRH theory more relevant for LLM interpretability, we must generalize \Cref{sec:linear-decomposition} to words made of multiple tokens. At first glance, the insight is simple: \textit{tokens do not build concepts -- tokens build words -- and words build concepts}. Nonetheless, words are sequences of tokens with a well defined order, implying we cannot simply average them to a single vector or their meaning might be lost, as exemplified at \Cref{tab:word-meaning}.

\begin{table}[tb] 
    \footnotesize
    \centering
    \begin{tabular}{cc}
        {\textbf{Synset}}       & {\textbf{Lemma}}  \\ \hline
        \synsetad{}              & \tokenad{}    \\
        \synsetmit{}            & \tokenmit{}   \\
        \synsetadmit{}          & \wordadmit{} \\
        \synsetmitad{}          & \wordmitad{} \\
    \end{tabular}
    \caption{An example of a token pair -- \tokenad{} and \tokenmit{} -- being used to form different words, each portraying different concepts. The Gemma 2 model family~\cite{team2024gemma} tokenizes \wordadmit{}, in the OMW \textit{synset} of \synsetadmit{}, into \tokenad{} in \synsetad{}, and the Polish word \tokenmit{} in \synsetmit{}. Concurrently, in opposite order they form the Spanish word \wordmitad{}, present in \synsetmitad{}. 
    }
    \label{tab:word-meaning}
\end{table}

\subsubsection{Words as Frames}

In practice, we define a word $\word{W}$ as an OMW lemma (\cf\Cref{sec:wordnet}). It is split into $\sentencelength$ independent tokens $(\tokensequence[\sentencelength]{w})$, and represented in $\unembeddingspace$ as a sequence of unembedding vectors, \ie, the matrix
\begin{equation}\label{def:word}
    \wordmat{} = \makematrix{\unembedding(\token{w}_1) & \unembedding(\token{w}_2) & \dots & \unembedding(\token{w}_\sentencelength)}.
\end{equation}

First, consider the following principle about the nature of word matrices. Let two token vectors $\vect{a}, \vect{b} \in \unembeddingspace$. If these vectors were collinear, meaning $\vect{b} = \alpha \vect{a}$ for some $\alpha \in \reals$, then \Cref{eq:llm-prob} would consistently assign higher probabilities to one token over the other regardless of input. This would effectively make some tokens redundant, as they would never be the most probable choice in any context. Such a scenario contradicts a fundamental design of LLMs, where each token in $\vocabulary$ must have some context in which it is the optimal choice, \ie, all tokens must be meaningful and usable. Thus, we conclude no two token vectors are collinear.

In that sense, word matrices as in \Cref{def:word} are constrained so that no token vector $\unembedding(\token{w}_i)$ can be expressed as a scalar multiple $\alpha \unembedding(\token{w}_j)$ of another token vector, for any $\alpha \in \reals$. This non-collinearity constraint defines a locally Euclidean open subset of $\reals^{\dimension \times \sentencelength}$, thereby forming a manifold. The softmax operation in DNN training ensures the space of all words acquires a manifold structure.

While non-collinearity is a necessary condition, it does not enforce $\wordmat{}$ as full-rank -- the $\sentencelength$ token vectors may exist in a subspace of dimension less than $\sentencelength$. However, we assume rank deficiency may compromise expressiveness and computational stability, suggesting the need for additional constraints.

From an NLP perspective, unique word representations are essential. If a word was not a linearly independent matrix, we could eliminate dependent tokens until made full-rank, yielding an alternative representation of the same word in $\unembeddingspace$. However, this sacrifices word uniqueness, which is undesirable for consistent language modeling.

To address these concerns, we propose modeling words as frames, \ie, we assume $\wordmat{} \in \Stiefel(\sentencelength, \dimension)$. Then, all word matrices are supposed full-rank. Our empirical investigation (\cf\Cref{sec:exp-frame-repr-hypothesis}) supports this framework, revealing that over 99\% of words in OMW exhibit linear independence among their token vectors. This assumption is facilitated by the high dimensionality of $\unembeddingspace$, easily representing words as full-rank matrices.




\subsubsection{Frame Correlation}\label{sec:frame-inner-product}

We name the set of all words the \spacename, or equivalently, $\completestiefelmanifold{\numbers}$, $\numbers$ the max number of tokens in any word. Let $\mat{A} = \makematrix{\vect{a}_1 & \dots & \vect{a}_{\numbers_1}} \in \stiefelmanifold{1}, \mat{B} = \makematrix{\vect{b}_1 & \dots & \vect{b}_{\numbers_2}} \in \stiefelmanifold{2}$ be frames of $\completestiefelmanifold{\numbers}$, we employ the asymmetric Procrustes distance~\cite{ye2016schubert,mandolesi2022asymmetric} as the space metric,
\begin{equation} \label{eq:distance-llm-space}
    \distcfasym(\mat{A},\mat{B}) = \sqrt{\numbers_1 + \numbers_2 - 2\sum_j^{\min \numbers_1,\numbers_2} \vect{a}_{j} \innerproductmatrix \vect{b}_{j}},
\end{equation}
where $\innerproductmatrix$ comes from \Cref{eq:lrh-inner-product}.

Hereafter, we can propose frame correlation by applying the law of cosines to generalize \Cref{eq:ray-corr} only in terms of distance functions:
\begin{align} \label{eq:frame-correlation}
    \corr(\mat{A},\mat{B}) 
    &= \frac{\norm{\mat{A}}_{\Procrustes}^2 + \norm{\mat{B}}_{\Procrustes}^2 - \distcfasym(\mat{A},\mat{B})^2}{2\norm{\mat{A}}_{\Procrustes}\norm{\mat{B}}_{\Procrustes}} \\
    &= \frac{\sum_j^{\min \numbers_1,\numbers_2} \vect{a}_{j}\innerproductmatrix \vect{b}_{j}}{\sqrt{\numbers_1 \numbers_2}},
\end{align}
where $\norm{\mat{A}}_{\Procrustes} = \distcfasym(\mat{A},\varnothing) = \sqrt{\numbers_1}$, $\varnothing$ is the null frame (origin) of $\completestiefelmanifold{\numbers}$, so $\rank(\varnothing) = 0$.

Such correlation can measure relationships as similar (positive), unrelated (null), or opposite (negative). For instance, \word{\green{yeah}} and \word{\green{yes}} are similar words and should have correlation close to 1, while \word{\green{yes}} and \word{\coolblue{bubble}} are orthogonal, but antonyms such as \word{\green{yes}} and \word{\purplepink{no}} would be negatively correlated. 

\subsubsection{Concept Frame}

We estimate concepts as the Fréchet mean of a word set -- the point minimizing the distance to each word -- effectively capturing the concept they collectively represent~\cite{marrinan2014finding}.

Let $\setwithcount{\wordmat{}}{i}$ be a set of words, $\wordmatdef{i} \in \stiefelmanifold{i}$, and let $\conceptmatdef \in \Stiefel(\numbers, \dimension), \numbers = \max \numbers_i$ be the Concept Frame, it is determined as
\begin{align}
    \conceptmat{} 
    \label{eq:word-frame-mean-1}
    &= \argmin_{\conceptmat{} \in \stiefelmanifold{}} \sum_{i = 1}^{\setsize} {\distcfasym}^2(\wordmat{i}, \conceptmat{}) \\
    \label{eq:word-frame-mean-2}
    &= \argmax_{\conceptvect{j} \in \Stiefel(1, \dimension)} \sum_{i = 1}^{\setsize} \sum_{j = 1}^{\numbers_i} \unembedding(\token{w}_{ij})^\top \innerproductmatrix \conceptvect{j}.  
\end{align}
We can extend the sum at \Cref{eq:word-frame-mean-2} from $\numbers_i$ to $\numbers$ by noticing its equivalence to having $\unembedding(\token{w}_{ij}) = 0$ for all $\numbers_i < j \le \numbers$. Let's define $\wordmat{i}' = \makematrix{\unembedding(\token{w}_{{i}1}) & \unembedding(\token{w}_{{i}2}) & \dots & \unembedding(\token{w}_{{i}\numbers_{{i}}}) & 0 & 0 & \dots & 0}$ as the right-padded $\wordmat{i}$ with $\numbers - \numbers_i$ zeros. Thus,
\begin{align} 
    \conceptmat{} &= \argmax_{\conceptvect{j} \in \Stiefel(1, \dimension)} \sum_{i = 1}^{\setsize} \sum_{j = 1}^{\numbers} {\wordmat{ij}'}^\top \innerproductmatrix \conceptvect{j} \\
    &= \argmax_{\conceptvect{j} \in \Stiefel(1, \dimension)} \sum_{j = 1}^{\numbers} (\sum_{i = 1}^{\setsize} {\wordmat{ij}'}^\top) \innerproductmatrix \conceptvect{j} \\
    \label{eq:procrustes-problem}
    &= \argmax_{\conceptmat{} \in \stiefelmanifold{}} \tr(\bar{\wordmat{}}'^\top \innerproductmatrix{} \conceptmat{}),
\end{align}
where $\bar{\wordmat{}}' = \sum_{i = 1}^{\setsize} \wordmat{i}'$ is the padded word sum.

Finally, \Cref{eq:procrustes-problem} is the Procrustes problem, which \citet{schonemann1966generalized} has solved with
\begin{equation}\label{eq:procustes-solution-frame-mean}
    \conceptmat{} = \mat{U} \mat{V}^\top,
\end{equation}
and $\bar{\wordmat{}}'^\top \innerproductmatrix{} = \mat{U} \mat{\Sigma} \mat{V}^\top$ is the SVD decomposition of the padded word sum. Hence, under the Procrustes distance the Concept Frame is the solution of a Procrustes problem.

\subsubsection{Combined Concept Frames}

In \Cref{sec:combined-concepts}, we defined Combined Concepts as concept vector differences, which we extend to \methodabbr by placing them in the Stiefel manifold. In other words, given a pair of Concept Frames $\mat{A}, \mat{B} \in \stiefelmanifold{}$, we can build the Combined Concept Frame $\diffframe = \diffframe(\mat{B},\mat{A}) \in \stiefelmanifold{}$ by enforcing it to be the frame closest to $\mat{B} - \mat{A}$:
\begin{equation}
    \diffframe = \argmax_{\diffframe \in \stiefelmanifold{}} \tr((\mat{B} - \mat{A})^\top \innerproductmatrix{} \diffframe),
\end{equation}
Thereby, $\diffframe(\mat{B},\mat{A}) = \mat{U}_{\diffframe} \mat{V}^\top_{\diffframe}$, $(\mat{B} - \mat{A})^\top \innerproductmatrix = \mat{U}_{\diffframe} \mat{\Sigma}_{\diffframe} \mat{V}^\top_{\diffframe}$ the SVD decomposition.



\subsection{Concept Probing}

The framework established for $\unembeddingspace$ can be extended to the feature space $\inputfeaturespace$ by reinterpreting \Cref{eq:llm-prob} as $\logit p(\token{y}|x) = \corr(\unembedding(\token{y}),\feature(\mathnormal{x}))$. Thus, the correlation between $\unembedding(\token{y})$ and $\feature(x)$ can be understood as a linear probe from space $\unembeddingspace$ to $\inputfeaturespace$.

Consequently, there is a correspondence between frames in $\unembeddingspace$ and $\inputfeaturespace$. Let a Feature Frame $\featuremat{}$ be the last $\numbers$ feature vectors of the input sequence $$\featuremat(\mathnormal{x}) = \makematrix{\feature_{\sentencelength - \numbers + \mathnormal{1}} & \feature_{\sentencelength - \numbers + \mathnormal{2}} & \dots & \feature_{\sentencelength}} \in \stiefelmanifold{},$$ we probe $x$ for Concept Frame $\conceptmat{} \in \stiefelmanifold{}$ using the correlation defined at \Cref{eq:frame-correlation}, 
\begin{equation}\label{eq:concept-probe}
    \logit p(\conceptmat{}|x) = \corr(\conceptmat{},\featuremat(\mathnormal{x})).
\end{equation}

\subsection{\guideddecoding{}}

We can leverage concept probing as a mechanism for Concept-Guided Text Generation (\Cref{fig:top_k_guided_decoding}). This approach can be implemented with sample-based decoding methods, such as Top-$k$ sampling, first generating a set of $k$ potential tokens from which the next token is randomly selected. We propose to alter such process wherein the next token $\token{x}_{t + 1}$ of input sequence $x = (\tokensequence{x})$ is the one which maximizes its respective Feature Frame correlation onto a target Concept Frame $\conceptmat{}$,
\begin{equation}
    \token{x}_{t + 1} = \argmax_{i \in \set{1, 2, \dots, k}} \corr(\conceptmat{},\featuremat_{\mathnormal{i}}(\mathnormal{x})).
\end{equation}

This methodology can align model output with a desired concept and serves as a practical prototype for \methodabbr, showing how to direct text generation and understand model behavior meaningfully.

\section{Experiments}

In this section, we validate \methodabbr for words and concepts, showing guided generation of sentences. We use Llama 3.1~\cite{llama31meta2024}, Gemma 2~\cite{team2024gemma}, and Phi 3~\cite{Phi3microsoft2024} LLM families and OMW only with supported languages. Further discussion is available in the Appendices. 

\subsection{\methodname}~\label{sec:exp-frame-repr-hypothesis}

\begin{figure}[tb]
    \centering
    \includegraphics[width=\linewidth]{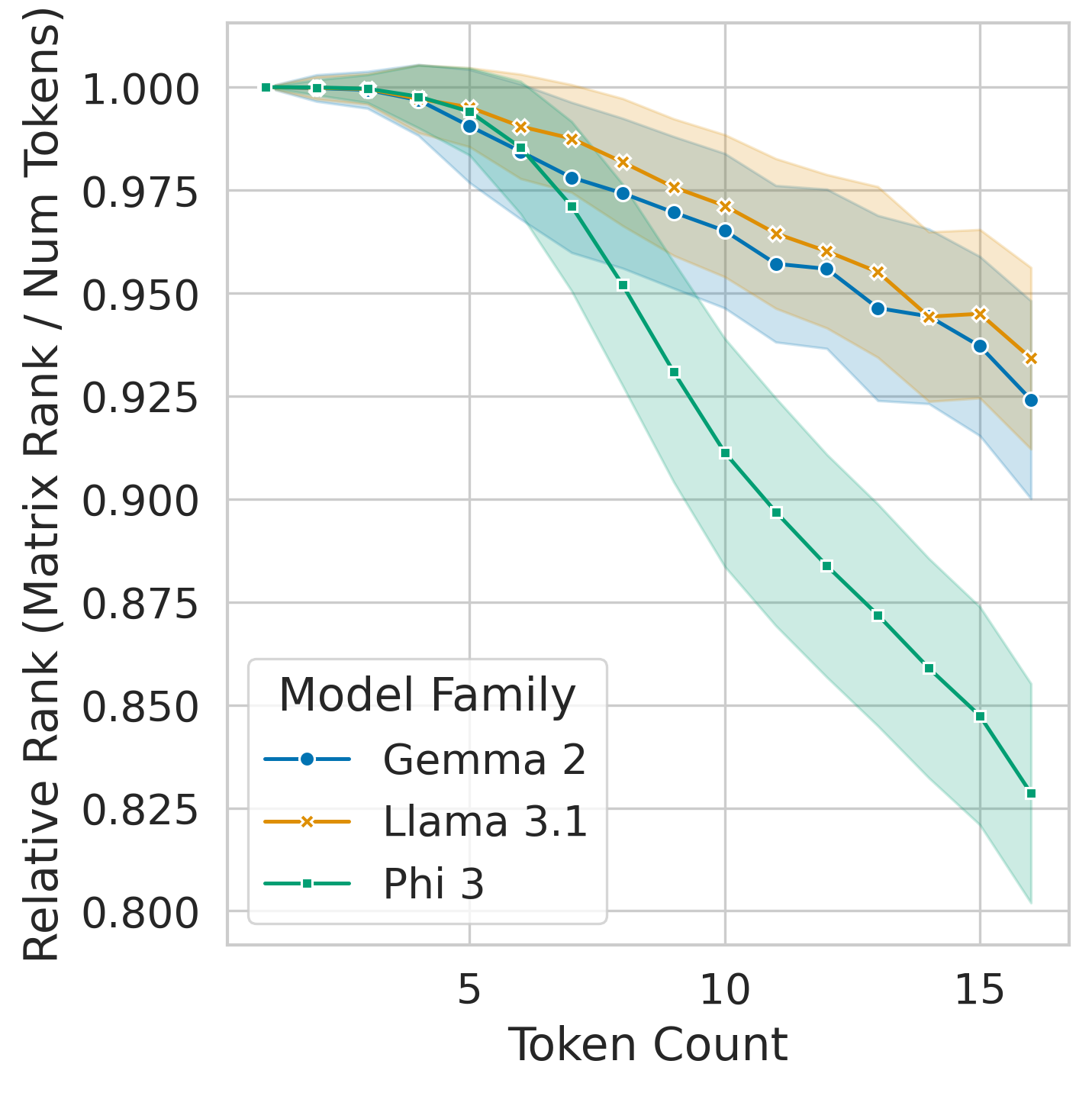}
    \caption{Relative Rank as a function of token count for all OMW lemmas and model families. Over 99\% of words are full-rank. Phi 3 has lower overall rank for longer lemmas than other models.}
    \label{fig:relative-rank-vs-token-count}
\end{figure}

\methodabbr posits LLMs encode words as frames. We can empirically evaluate this hypothesis by analyzing if words are made of linearly independent vectors, which we can measure by computing its rank. In \Cref{fig:relative-rank-vs-token-count}, we see near-maximum matrix ranks for lemmas comprising up to 3-4 tokens, which is the token count that represents words. In OMW, lemmas with token counts of 5 and beyond mostly represent compound words and expressions, implying the frame representation fits $99.8\%$ words. Notably, Phi 3 shows a rapid rank decrease beyond token count of 5, likely due to its high proportion of lemmas with large token count, making non full-rank lemmas more common (\cf \Cref{fig:hist-token-count}).

\begin{figure}[tb]
    \centering
    \includegraphics[width=\linewidth]{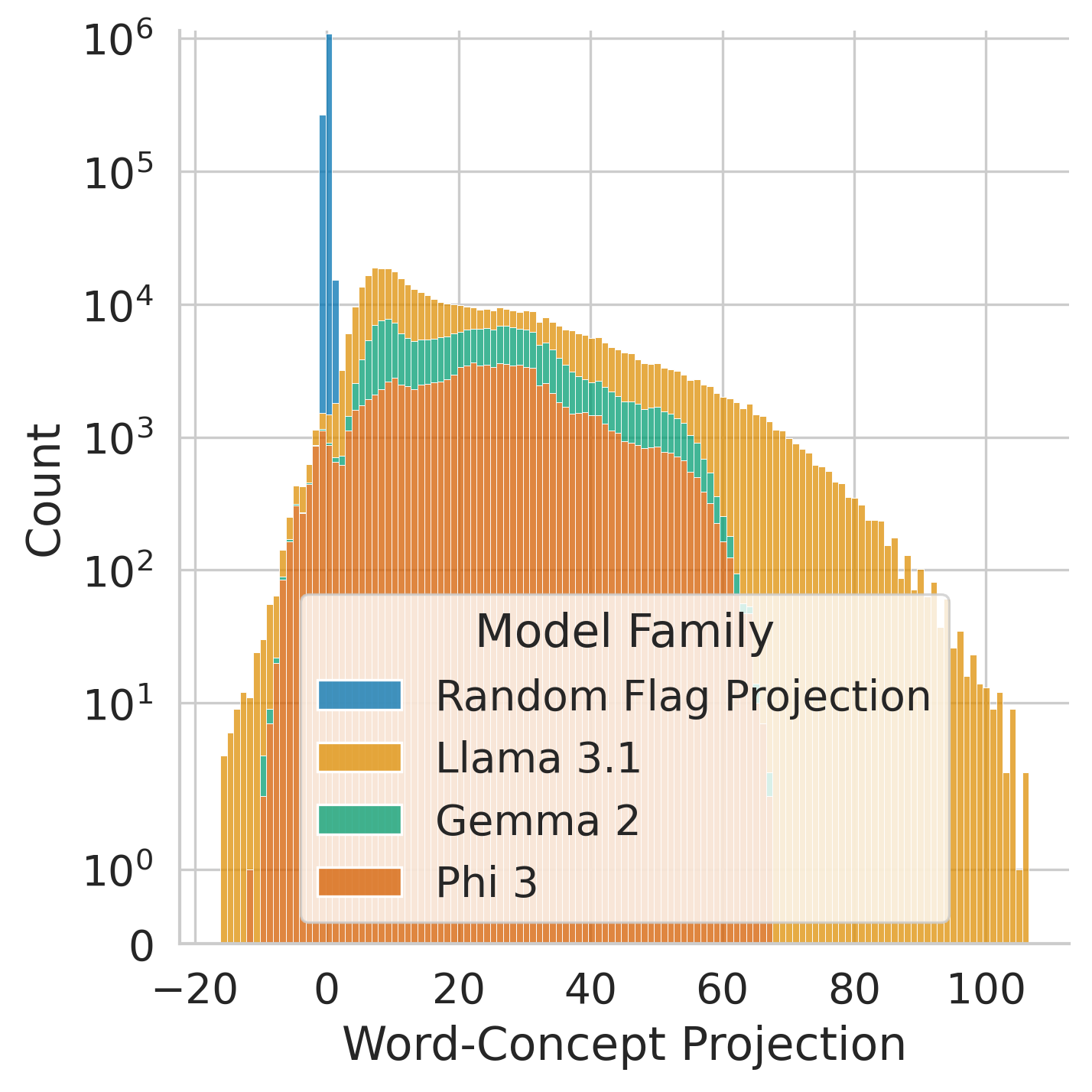}
    \caption{Distribution of word frame projection lengths. Random frames have near-zero projection with any Concept Frame, while words show positive projections onto associated Concept Frames.}
    \label{fig:concept_vs_word_frame_relationship}
\end{figure}

Furthermore, given we propose using OMW synsets to build Concept Frames, we must verify if these synsets fit the model representation or not. To that end, we can compute the projection (unnormalized correlation) of Word Frames onto their corresponding Concept Frames for all OMW synsets and lemmas. \Cref{fig:concept_vs_word_frame_relationship} reveals that random frames are consistently unrelated to concept frames across models, while words exhibit positive projections onto their associated concepts.

These findings support \methodabbr consistency with models' internal representations and suggest LLMs inherently correlate with the OMW linguistic graph. In the following experiments, we use lemmas up to 4 tokens to ensure our theory is applied only for full-rank matrices.

\subsection{Guided Generation}

Given the \methodabbr evidences, we explore its application in text generation with \guideddecoding{}, exposing biases and vulnerabilities.

\subsubsection{Qualitative Analysis} \label{sec:guided_generation_qualitative}

We first compare model outputs on a few inputs and concepts likely sensitive to biases. The example at \Cref{tab:qualitative-men} demonstrates the impact of concept-guided text generation on the characterization of \word{men} by Llama 3.1 8B Instruct. With no guidance, the model focuses on family roles. When guided by the Concept Frame $\token{woman.n.01 - male.n.01}$, this tendency is seemingly amplified. However, a more significant shift in narrative occurs when the model is guided by the opposite concept of $\token{male.n.01 - woman.n.01}$, prompting it to emphasize a perceived importance as \textit{family providers}.

\begin{figure}[!h]
    \user{What men can be?}

    \assistant{Men can be fathers, sons, brothers, and husbands. 
    }{no guidance}
    
    \assistant{1. A husband. 2. A father. 3. A son. 4. A friend. 5. A boyfriend or partner...}{$\token{woman.n.01 - male.n.01}$}
    
    \assistant{Men as fathers and family providers, as well as caregivers, are essential for family well-being...}{$\token{male.n.01 - woman.n.01}$}
    
    \caption{Concept-guided generation examples when the model is prompted to describe \word{\green{men}}.}
    \label{tab:qualitative-men}
\end{figure}

\begin{figure}[!h]
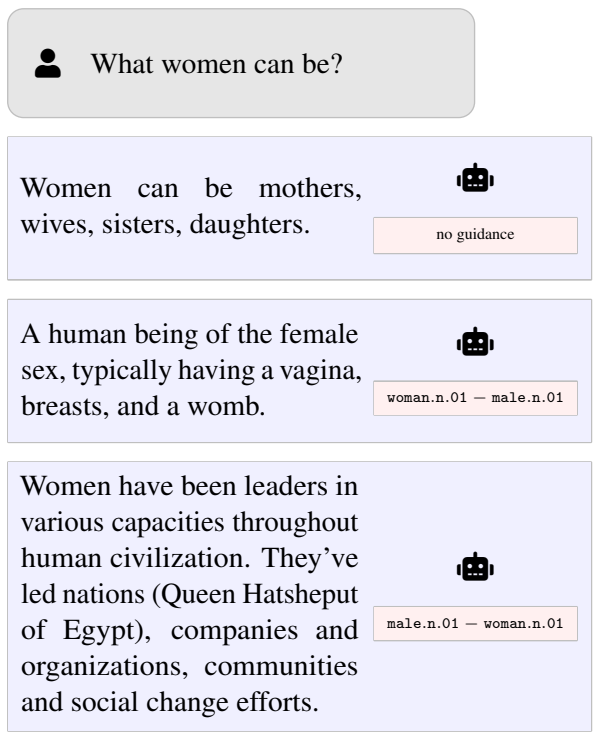

    \user{What women can be?}
    
    \assistant{Women can be mothers, wives, sisters, daughters.}{no guidance}
    
    \assistant{A human being of the female sex, typically having a vagina, breasts, and a womb.}{$\token{woman.n.01 - male.n.01}$}
    
    \assistant{Women have been leaders in various capacities throughout human civilization. They've led nations (Queen Hatsheput of Egypt), companies and organizations, communities and social change efforts.}{$\token{male.n.01 - woman.n.01}$}
    
    \caption{Concept-guided generation examples when the model is prompted to describe \word{\green{women}}.}
    \label{tab:qualitative-women}
\end{figure}

At \Cref{tab:qualitative-women}, the unguided model's characterization of \word{women} primarily enumerates family roles, which may be interpreted as a balanced output when juxtaposed with the default answer for \word{men} at \Cref{tab:qualitative-men}. Despite that, when guided by $\token{woman.n.01 - male.n.01}$, the model's output noticeably emphasizes biological traits. Conversely, $\token{male.n.01 - woman.n.01}$ highlights leadership roles and esteemed social positions.

The stark contrast in each example suggests that, when guided by a Combined Concept $\diffframe(\concept{B},\concept{A})$, the model attempts to maximize attributes it associates with the first concept $\concept{B}$ while minimizing the second concept $\concept{A}$. They illustrate how to influence text generation, exposing biases and stereotypes within the model's learned representations.

Notably, most generations kept a high level of readability, but using elevated values of $k$ can lead to incoherent text, a known issue of top-$k$ sampling~\cite{Holtzman2019TheCC}.

Besides, this process can expose vulnerabilities, including the capacity to generate harmful content, exemplified in \Cref{app:text-completion-examples}. The authors emphatically discourage this tool usage for malicious purposes yet acknowledge its potential for misuse, but more studies are warranted to comprehend their extent and implications.

\subsubsection{Quantitative Analysis}

A comprehensive understanding begets a quantitative study. We used a multilingual instruction dataset to ensure a minimum of 1000 sentences for each model supported language. The concept of choice was $\token{woman.n.01 - man.n.01}$ to stay consistent with the previous section. Resource constraints limited our investigation to a single concept, though we argue the results are indicative of the model's behavior across similar conceptual domains.

\begin{figure}[tb]
    \centering
    \includegraphics[width=\linewidth]{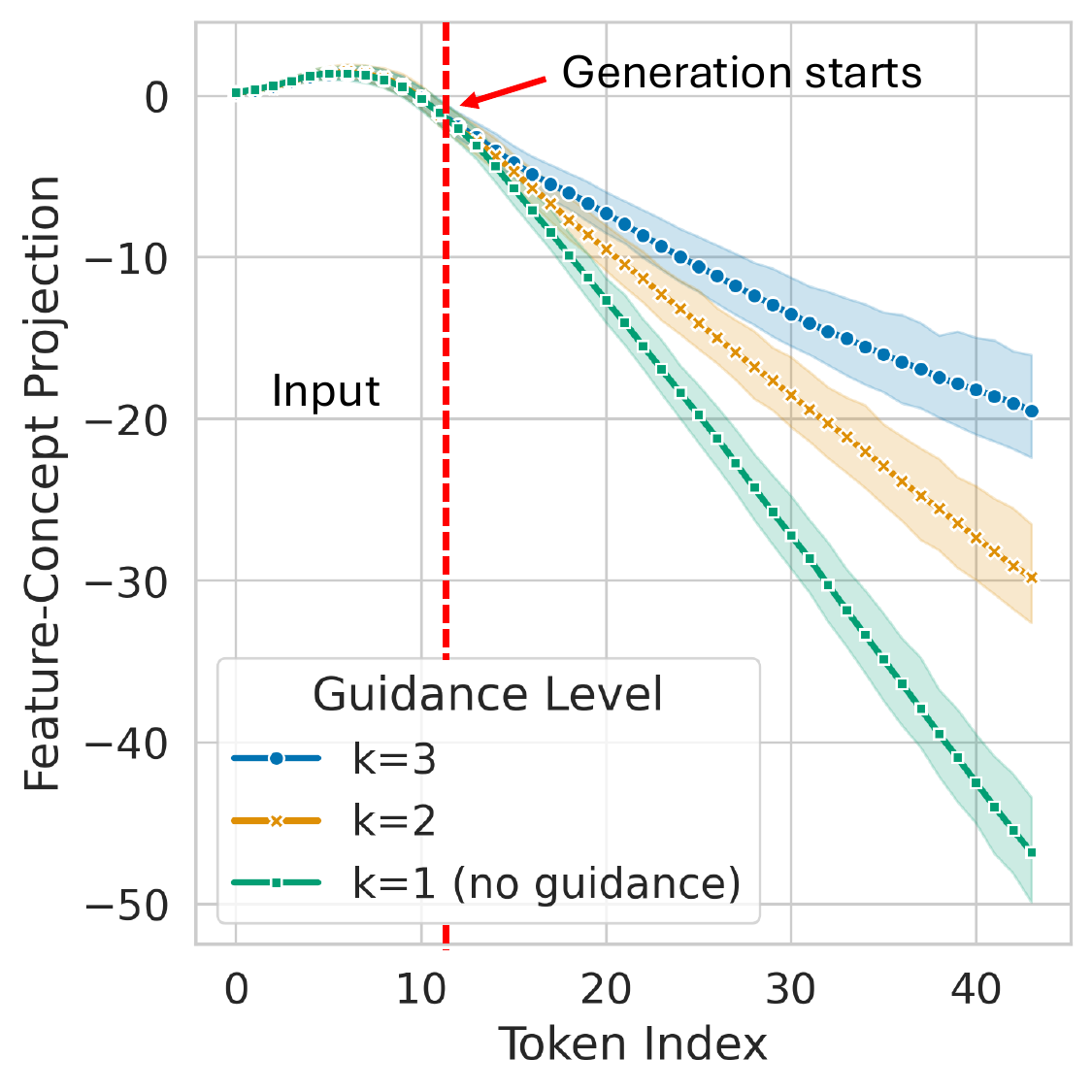}
    \caption{Concept probing evolution for 3 levels of \guideddecoding{} with Llama 3.1 70B AWQ~\cite{Lin2023AWQAW}. The guidance with $\token{woman.n.01 - man.n.01}$ is able to counter the LLM tendency to maximize $\token{man.n.01}$.}
    \label{fig:guided_generation_topk_comparison}
\end{figure}

Initially, we focused on the evolution of generated sentences across distinct values of $k$. As visible in \Cref{fig:guided_generation_topk_comparison}, all sentences start with minimal correlation to the chosen concept, evidenced by near-zero projection length. Notably, the unguided output naturally minimizes the projection with our chosen concept, indicating it tends toward the opposite direction of $\token{man.n.01 - woman.n.01}$. However, the algorithm demonstrated capacity to steer the output toward the desired concept with increasing effectiveness as $k$ increased, showing $k$ can regulate guidance strength. We highlight this result indicates biases in standard generation, and while guidance does not completely modify this scenario, it is remediated to a certain extent.

\begin{figure}[tb]
    \centering
    \includegraphics[width=\linewidth]{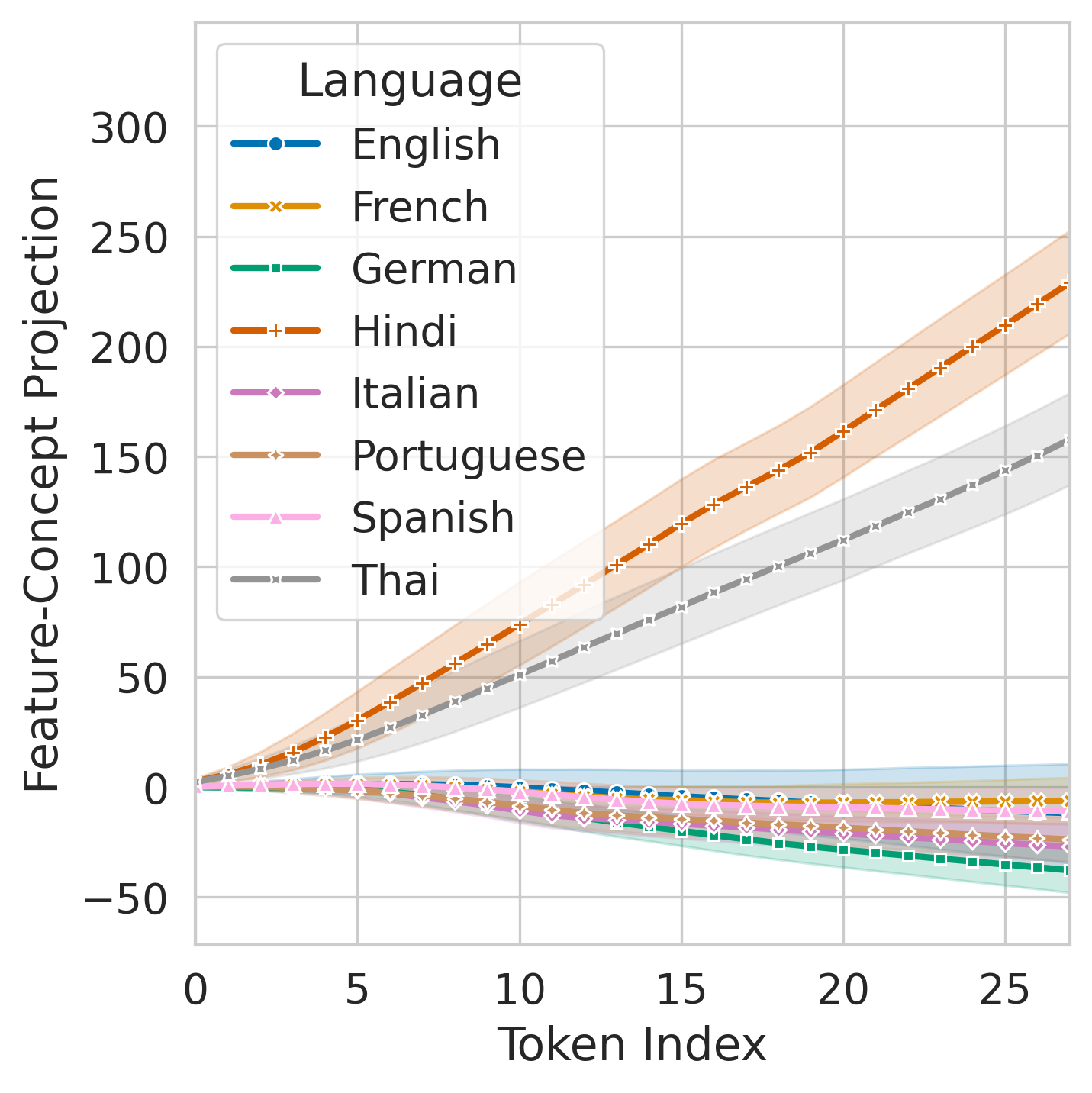}
    \caption{Concept probing evolution during model generation for the 8 languages supported by Llama 3.1 70B using \guideddecoding{} with $k = 3$. Hindi and Thai are more susceptible to the technique than other languages.}
    \label{fig:guided_generation_lang_comparison}
\end{figure}

Next, we examine concept-guided generation across Llama 3.1 supported languages. We find most languages exhibit comparable patterns, with Hindi and Thai serving as notable exceptions (\Cref{fig:guided_generation_lang_comparison}). These demonstrate significantly higher susceptibility to guidance and are the only non-european ones, suggesting the model treats said languages differently~\cite{llama31meta2024}. Further investigation is shown in \Cref{sec:guideddecoding-language-comparison}.

\begin{figure}[tb]
    \centering
    \includegraphics[width=\linewidth]{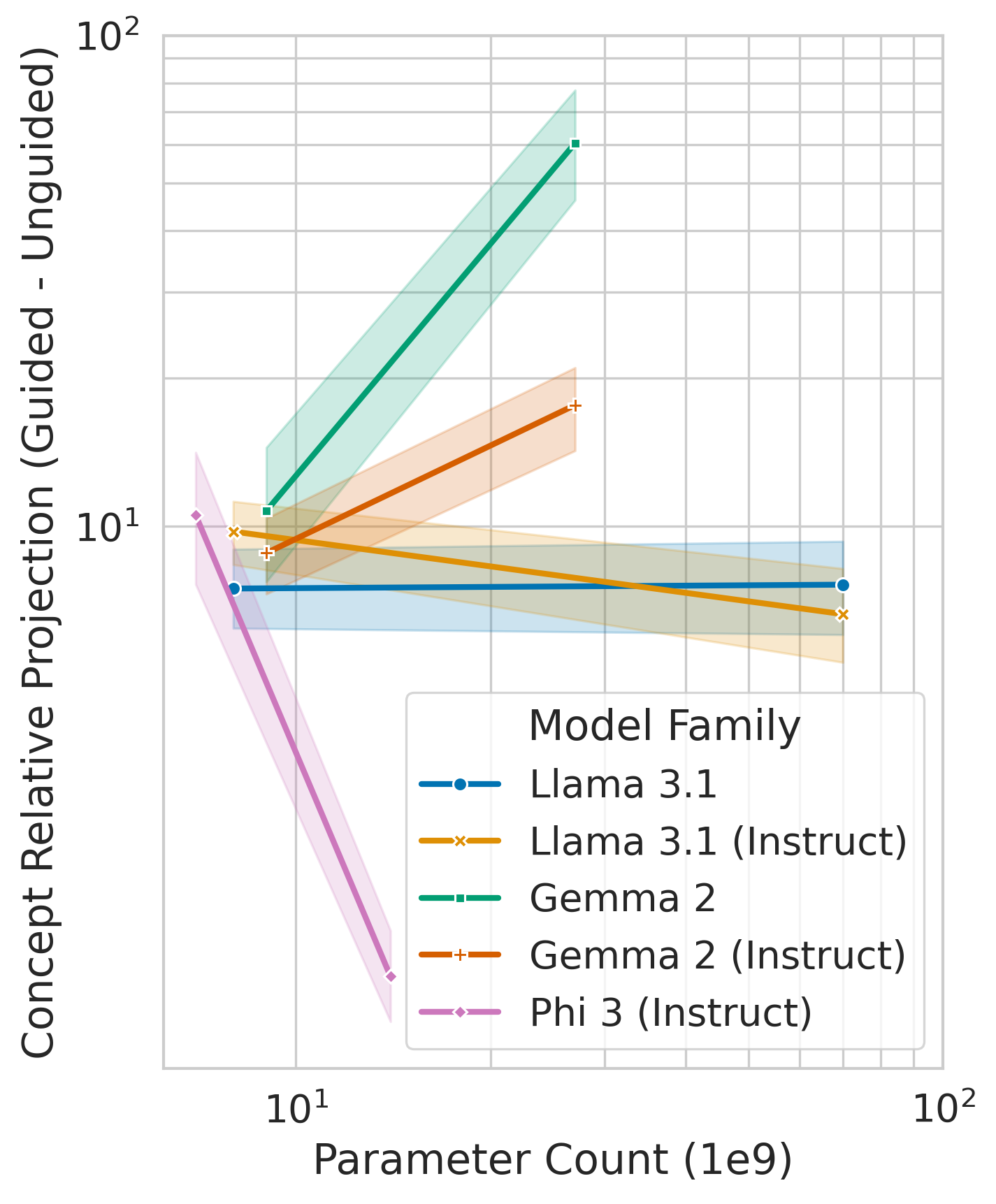}
    \caption{Concept relative projection for several models and parameter counts.}
    \label{fig:guided_generation_model_comparison}
\end{figure}

Finally, in \Cref{fig:guided_generation_model_comparison} we use the concept relative projection -- difference of guided and unguided projection to the concept -- to measure guidance susceptibility among several model families and various parameter counts. Llama 3.1 models seem equally susceptible to guidance among base and instruct models for all parameter counts; Gemma 2 shows more susceptibility to guidance when the parameter count increases, but there is a sensible reduction from base to instruct variations; On the other hand, Phi shows a great reduction in guidance effect with parameter count, possibly an effect of a less linear feature space as previously commented. Most curiously, guidance susceptibility is almost equivalent on all models for the lower parameter count, which could indicate a common convergence of representations.

\section{Conclusions}

This study proposes the \methodname, an extension of the Linear Representation Hypothesis with Lie Group elements. \methodabbr posits LLMs encode words as frames, with model input and output connected as Concept Frames in Stiefel Manifolds. \methodabbr provides a structured framework for LLM interpretability and control via concept probing and concept-guided decoding, showing that even state-of-the-art LLMs exhibit gender and language biases or harmful vulnerabilities.

This work is an initial exploration, and further research is made necessary to understand its extents. In particular, we have yet to explore $2^{nd}$ order Combined Concepts and higher, which could reveal even richer concept relationships, uncovering LLMs own ontology. In that sense, our concepts were limited to WordNet selection of meanings, and while it enabled efficient concept extraction without additional training, future work should integrate \methodabbr with Dictionary Learning techniques to automatically extract concepts from the model weights. Also, \guideddecoding{} served as a \methodabbr proof-of-concept, but is limited by the same constraints as Top-$k$ sampling, so we encourage more advanced and custom variations.

In conclusion, \methodabbr represents a promising avenue for LLM interpretability, and could lead to novel developments in safe, trustworthy and reliable AI systems.






   



  

\bibliographystyle{acl_natbib}
\bibliography{zotero}

\clearpage
\appendix

\section{Mathematical Details and Proofs} \label{sec:proofs}

\subsection{Linear Decomposition of Tokens}

\begin{lemma}[Concept estimation]
    Let $\setwithcount{\token{y}}{j}$ be a set of tokens sharing a common concept $\conceptvect{}$, we can estimate the concept as

    \begin{equation}
        \conceptvect{} \appropto \setsum{j} \unembedding(\token{y}_j) - \unembedding_0,
    \end{equation}
    with error of order $\mathcal{O}(\frac{1}{\sqrt{\setsize}})$.

    \begin{proof} \label{lemma:concept-estimation-proof}
        Let $\unembedding_j = \unembedding(\token{y}_j)$.
        If $\setwithcount{\unembedding}{j}$ share a common meaning $\conceptvect{}$, then by \Cref{ax:lin-comb-concept} every $\unembedding_j$ is represented as
        \begin{equation}
            \unembedding_j - \unembedding_0 = \sum_{i}^{\conceptcount} a_i \conceptvect{i} = \conceptcoef{j} \conceptvect{} + \sum_{\conceptvect{i} \neq \conceptvect{}}^n a_i \conceptvect{i} 
        \end{equation}
        where $\conceptcoef{j}$ is the coefficient of $\conceptvect{}$ for the token $\unembedding_j$. Then, 
        \begin{align}
            \setsum{j} \unembedding_j - \unembedding_0 &= \setsum{j} (\conceptcoef{j} \conceptvect{} + \sum_{\conceptvect{i} \neq \conceptvect{}} a_i \conceptvect{i}) 
            \\ 
            &= \setsum{j} \conceptcoef{j} \conceptvect{} + \setsum{j} \sum_{\conceptvect{i} \neq \conceptvect{}}^\setsize a_i \conceptvect{i} 
            \\ 
            &= \conceptcoef{} \conceptvect{} + \setsum{j} \sum_{\conceptvect{i} \neq \conceptvect{}}^\setsize a_i \conceptvect{i} 
            \\
            &= \conceptcoef{} \conceptvect{} + \mathcal{O}(\frac{1}{\sqrt{\setsize}})
        \end{align}
        where $\conceptcoef{} = \setsum{j} \conceptcoef{j}$ and $\setsum{j} \sum_{\conceptvect{i} \neq \conceptvect{}}^\setsize a_i \conceptvect{i}$ is a rough estimate of the sample mean, which should tend to zero with error equal to the standard error of the mean (SEM), considering the common concept factors more distinctly than the others.
    \end{proof}
\end{lemma}

\begin{proposition}[1st-order Concepts]
    A concept ${\concept{C}}$ has ray representation $\ray(\conceptvect{{\concept{C}}}')$, where $\conceptvect{{\concept{C}}}'$ is a normalized counterfactual concept pair difference:
    
    \begin{equation}\label{eq:1st-order-concept}
        \conceptvect{{\concept{C}}}' = \frac{\conceptvect{{\concept{C}}=1} - \conceptvect{{\concept{C}}=0}}{\norm{\conceptvect{{\concept{C}}=1} - \conceptvect{{\concept{C}}=0}}}
    \end{equation}

    \begin{proof} \label{prop:concept-der-proof}
        Following \Cref{eq:un-repr-2}, the unembedding representation of a concept ${\concept{C}}$ is computed as a normalized mean of counterfactual pairs. Thus,

        \begin{align}
            \unembedding_{\concept{C}}' &= \sum_{i}^{n_{\concept{C}}} (\unembedding_i({\concept{C}}=1) - \unembedding_i({\concept{C}}=0)) \\
            &= \sum_{i}^{n_{\concept{C}}} \unembedding_i({\concept{C}}=1) - \sum_{i}^{n_{\concept{C}}} \unembedding_i({\concept{C}}=0) \\
            &= \sum_{i}^{n_{\concept{C}}} (\unembedding_i({\concept{C}}=1) - \unembedding_0) \\ &- \sum_{i}^{n_{\concept{C}}} (\unembedding_i({\concept{C}}=0) - \unembedding_0) \\
            &= \conceptvect{{\concept{C}}=1} - \conceptvect{{\concept{C}}=0}
        \end{align}
        where $\conceptvect{{\concept{C}}=1} = \sum_{i}^{n_{\concept{C}}} (\unembedding_i({\concept{C}}=1) - \unembedding_0)$ and $\conceptvect{{\concept{C}}=0} = \sum_{i}^{n_{\concept{C}}} (\unembedding_i({\concept{C}}=0) - \unembedding_0)$ are concepts for each counterfactual pair item. Then, normalizing $\unembedding_{\concept{C}}'$ gives $\conceptvect{{\concept{C}}}'$ at \Cref{eq:1st-order-concept}.
    \end{proof}
\end{proposition}

\subsection{\methodname}

\subsubsection{Rays and Subspaces}

Let $\vect{v}, \vect{u} \in \vectorspace$ be two vectors, angle $\theta$ between them, their respective rays and 1-dim subspaces are two distinct structures which can be generalized to points in Grassmann manifolds differing only by choice of distance. Rays use the chordal Frobenius distance, also known as the Procrustes distance~\cite{mandolesi2022asymmetric}, given by
\begin{equation}
    \distcf(\ray(\vect{v}), \ray(\vect{u})) = \norm{\vect{v} - \vect{u}}_{\Frobenius} = 2 \sin \frac{\theta}{2},
\end{equation}
where $\norm{\cdot}_{\Frobenius}$ is the Frobenius norm, making the Frobenius inner product the space inner product.

In this context, correlation (\ref{eq:ray-corr}) is induced by the choice of distance and norm:
\begin{align}
    \corr(\ray(\vect{v}), \ray(\vect{u})) 
    &= \frac{\innerproduct{\vect{v}}{\vect{u}}_{\Frobenius}}{\norm{\vect{v}}_{\Frobenius}\norm{\vect{u}}_{\Frobenius}} \\
    &= \cos \theta
\end{align}
where we use the term ``correlation'' to indicate a generalization of the traditional \textit{cosine similarity} beyond just vectors.

On the other hand, their respective subspaces $\subspacemat{\vect{v}}, \subspacemat{\vect{u}}$ are compared using the projective distance
\begin{align}
    \dist_{\Projection}(\subspacemat{\vect{v}}, \subspacemat{\vect{u}}) &= \norm{\vect{v}\vect{v}^\top - \vect{u}\vect{u}^\top}_{\Frobenius} \\ 
    &= \sqrt{1 - \cos^2 \theta}.
\end{align}

Thus,
\begin{align}
    \corr(\subspacemat{\vect{v}}, \subspacemat{\vect{u}}) 
    &= \frac{\innerproduct{\vect{v}\vect{v}^\top}{\vect{u}\vect{u}^\top}_{\Frobenius}}{\norm{\vect{u}\vect{u}^\top}_{\Frobenius}\norm{\vect{u}\vect{u}^\top}_{\Frobenius}} \\
    &= \cos^2 \theta.
\end{align}

\subsubsection{Combined Concept Geometrical Interpretation}

We show a geometrical interpretation of the Combined Concept Frame, illustrated in \Cref{fig:difference-subspace-manifold}.

\begin{proposition}
    Let $\mat{A}, \mat{B} \in \stiefelmanifold{}$ be Concept Frames (\ref{eq:procustes-solution-frame-mean}). Then, $\diffframe(\mat{B},\mat{A})$ is the frame which best approximates the direction of the geodesic from $\mat{A}$ to $\mat{B}$ at its midpoint.

    \begin{proof}
        Let $\gamma(t) = \mat{A} \exp (t \mat{\Omega})$ be the geodesic connecting $\mat{A}$ and $\mat{B}$, $\mat{\Omega} = \log (\mat{A}^\top \mat{B}) \in \reals^{\numbers \times \numbers}$. Given the matrix exponential MacLaurin series $\exp(\mat{X}) = \sum_{n = 0}^{\infty} \mat{X}^n / n!$, we find that the derivative of the geodesic at the midpoint is
        \begin{align}
            \gamma'(\onehalf) &= \mat{A} \exp (\mat{\Omega}/2) \mat{\Omega} \\
            &= \mat{A} \sum_{n = 0}^\infty \frac{\mat{\Omega}^{n + 1}}{n! 2^n} \\
            &= \mat{A} (\mat{\Omega} + \frac{\mat{\Omega}^2}{2} + \dots)
        \end{align}
        
        Similarly,
        \begin{align}
            \mat{B} - \mat{A} &= \mat{A}(\exp \mat{\Omega} - \mat{I}) \\
            &= \mat{A}\sum_{n = 1}^\infty \frac{\mat{\Omega}^n}{n!} \\
            &= \mat{A} (\mat{\Omega} + \frac{\mat{\Omega}^2}{2} + \dots)
        \end{align}
        
        The series match to second order. Therefore, if $\mat{A}$ and $\mat{B}$ are not unrelated concepts, $\mat{B} - \mat{A} \approx \gamma'(\onehalf)$. Since $\diffframe(\mat{B},\mat{A})$ is the closest frame to $\mat{B} - \mat{A}$, $\diffframe(\mat{B},\mat{A})$ simultaneously approximates the direction of the geodesic at the midpoint.
    \end{proof}
\end{proposition}

\begin{figure}[h]
    \centering 
    \begin{subfigure}{\textwidth}
        \begin{tikzpicture}[isometric view]
            \draw[ball color=blue!50] {[canvas is xy plane at z=0] (135:2) arc (135:315:2)} arc (0:180:2cm);
    
            \coordinate (O) at (0,0,0);
            \coordinate (A) at (-1,1,0.75);
            \coordinate (B) at (1,-1,0.75);
            \coordinate (M) at (0,0,1.8); 
    
            \draw[thick, white] plot[smooth, tension=1] coordinates {(A) (M) (B)};
    
            \filldraw[black] (A) circle (2pt) node[below right] {$\subspace{A}$};
            \filldraw[black] (B) circle (2pt) node[below left] {$\subspace{B}$};
            \filldraw[red] (M) circle (2pt) node[above] {$\subspace{M}$}; 
    
            \coordinate (A2) at ($(A) + (0.1,-0.1,0)$);
            \coordinate (B2) at ($(B) + (-0.1,0.1,0)$);
    
    
            \draw[thin, gray, fill=gray, fill opacity=0.5] (M) ++(-2.5,-2.5,0) -- ++(5,0,0) -- ++(0,5,0) -- ++(-5,0,0) -- cycle;
    
            \coordinate (A') at (-2, 1.5, 2);
            \coordinate (B') at (1.5, -2, 2.1);
    
            \filldraw[red] (A') circle (2pt) node[below] {$\log_{\subspace{M}}{\subspace{A}}$};
            \filldraw[red] (B') circle (2pt) node[below] {$\log_{\subspace{M}}{\subspace{B}}$};
    
            \coordinate (MA'') at ($(M) + (-0.1,0.1,0)$);
            \coordinate (A'') at ($(A') + (0.1,-0.1,0)$);
            
            \coordinate (MB'') at ($(M) + (0.1,-0.1,0)$);
            \coordinate (B'') at ($(B') + (-0.1,0.1,0)$);
    
            \draw[->, thick, purple] (MA'') -- (A'') node[midway, above] {$\subspace{A} - \subspace{B}$};
            \draw[->, thick, purple] (MB'') -- (B'') node[midway, above] {$\subspace{B} - \subspace{A}$};
        \end{tikzpicture}
    \end{subfigure}

    \vspace{1cm}
    
    \begin{subfigure}{\textwidth}
        \begin{tikzpicture}[isometric view]
            \draw[ball color=blue!50] {[canvas is xy plane at z=0] (135:2) arc (135:315:2)} arc (0:180:2cm);

            \coordinate (O) at (0,0,0);
            \coordinate (A) at (-1,1,0.75);
            \coordinate (B) at (1,-1,0.75);
            \coordinate (M) at (0,0,1.8); 

            \draw[thick, white] plot[smooth, tension=1] coordinates {(A) (M) (B)};

            \filldraw[black] (A) circle (2pt) node[below right] {$\subspace{A}$};
            \filldraw[black] (B) circle (2pt) node[below left] {$\subspace{B}$};
            \filldraw[red] (M) circle (2pt) node[above] {$\subspace{M}$}; 

            \coordinate (A2) at ($(A) + (0.1,-0.1,0)$);
            \coordinate (B2) at ($(B) + (-0.1,0.1,0)$);
            \coordinate (B') at (1.5, -2, 2.1);
            \coordinate (MB'') at ($(M) + (0.1,-0.1,0)$);
            \coordinate (B'') at ($(B') + (-0.1,0.1,0)$);

            \draw[->, thick, purple] (MA'') -- (A'') node[midway, above left] {$\gamma'_{\subspace{B}\rightarrow\subspace{A}}(t = \onehalf)$};
            \draw[->, thick, purple] (MB'') -- (B'') node[midway, above right] {$\gamma'_{\subspace{A}\rightarrow\subspace{B}}(t = \onehalf)$};
        \end{tikzpicture}
    \end{subfigure}

    \caption{Illustrations of Difference Frames on the Stiefel manifold. \textbf{Top:} Frames $\subspace{A}$ and $\subspace{B}$ are projected onto tangent space at the midpoint $\subspace{M}$ (frame mean). Difference subspaces are represented by arrows connecting $\subspace{M}$ to each Logarithm Map. \textbf{Bottom:} Difference Frames are approximately equivalent to the gradient of the geodesic between the two frames at the midpoint $\subspace{M} = \gamma_{\subspace{A}\rightarrow\subspace{B}}(\onehalf) = \gamma_{\subspace{B}\rightarrow\subspace{A}}(\onehalf)$, thereby $\subspace{B} - \subspace{A} \approx \gamma'_{\subspace{A}\rightarrow\subspace{B}}(\onehalf)$, while $\subspace{A} - \subspace{B} \approx \gamma'_{\subspace{B}\rightarrow\subspace{A}}(\onehalf)$}
    \label{fig:difference-subspace-manifold}
\end{figure}

\section{\guideddecoding{}} \label{sec:guideddecodingsamples}

\subsection{Text Completion Examples}
\label{app:text-completion-examples}
These are Text Completion samples from Llama 3.1 8B. We show extra examples of the guided decoding mechanism, including text completions and harmful instructions. The grey area is the input and the remaining text model output. 


\assistant{\highlight{men are} not allowed to wear skirts. A. true B. false Answer: B}{no guidance}

\assistant{\highlight{men are} not allowed to wear shorts in public. A. in Saudi Arabia. B. in Iran. C. anywhere. D. in the United States. Answer: A}{$\token{woman.n.01 - male.n.01}$} 

\assistant{\highlight{women are} from venus men are from mars.}{no guidance}

\assistant{\highlight{women are} not the only ones to suffer the effects of the patriachal society, but also the men.}{$\token{man.n.01 - woman.n.01}$} 

\assistant{\highlight{kids are} the future of the world.}{no guidance}

\assistant{\highlight{kids are} the future of our country and we need them in school to learn and be educated so that we have the best future for our country.}{$\token{man.n.01}$} 

\assistant{\highlight{children are} the future of the world.}{no guidance} 

\assistant{\highlight{children are} a gift. A friend sent me an article today where it {[}...{]} showed a link in a childless woman having an 8-9 year increase on her risk for breast cancer than a mother.}{$\token{woman.n.01}$} 

Text Completion samples from Gemma 2 9B:

\assistant{\highlight{men are} not the only ones who can be a victim of domestic violence.}{no guidance}

\assistant{\highlight{men are} the most important part in a woman's existence}{$\token{man.n.01}$}

\assistant{\highlight{women are} the most beautiful creatures on earth.}{no guidance}

\assistant{\highlight{women are} the most important part in the world.}{$\token{woman.n.01}$}

\assistant{\highlight{kids are} so cute.}{no guidance}

\assistant{\highlight{kids are} so cute and adorable, they can do anything to get your heart mel}{$\token{man.n.01 - woman.n.01}$}

\assistant{\highlight{children are} the future of the country.}{no guidance}

\assistant{\highlight{children are} a great blessing to us as a couple but also a great burden to the church.}{$\token{woman.n.01 - man.n.01}$}











\section{Extra Experimental Results} \label{sec:experiments-appendices}


\subsection{Experimental Settings} \label{sec:experimental-settings}

Experiments were conducted on a single Nvidia H100 80Gb GPU, so the largest Llama 3.1 405B was not considered. The multilingual instruction dataset was compiled from questions sourced from the Aya Dataset~\cite{singh2024aya}, supplemented with additional samples in Italian from SQuAD Italian~\cite{Croce2018NeuralLF}, German and Thai from XQuAD~\cite{Artetxe2019OnTC}, and Hindi from MLQA~\cite{Lewis2019MLQAEC}, ensuring a minimum of 1000 sentences for each of the 8 languages supported by Llama 3.1 models. Gemma 2 and Phi were restricted to English samples, following their official language support.

\subsection{\guideddecoding{} Language Comparison} \label{sec:guideddecoding-language-comparison}

\begin{figure}[!ht]
    \centering
    \includegraphics[width=\linewidth]{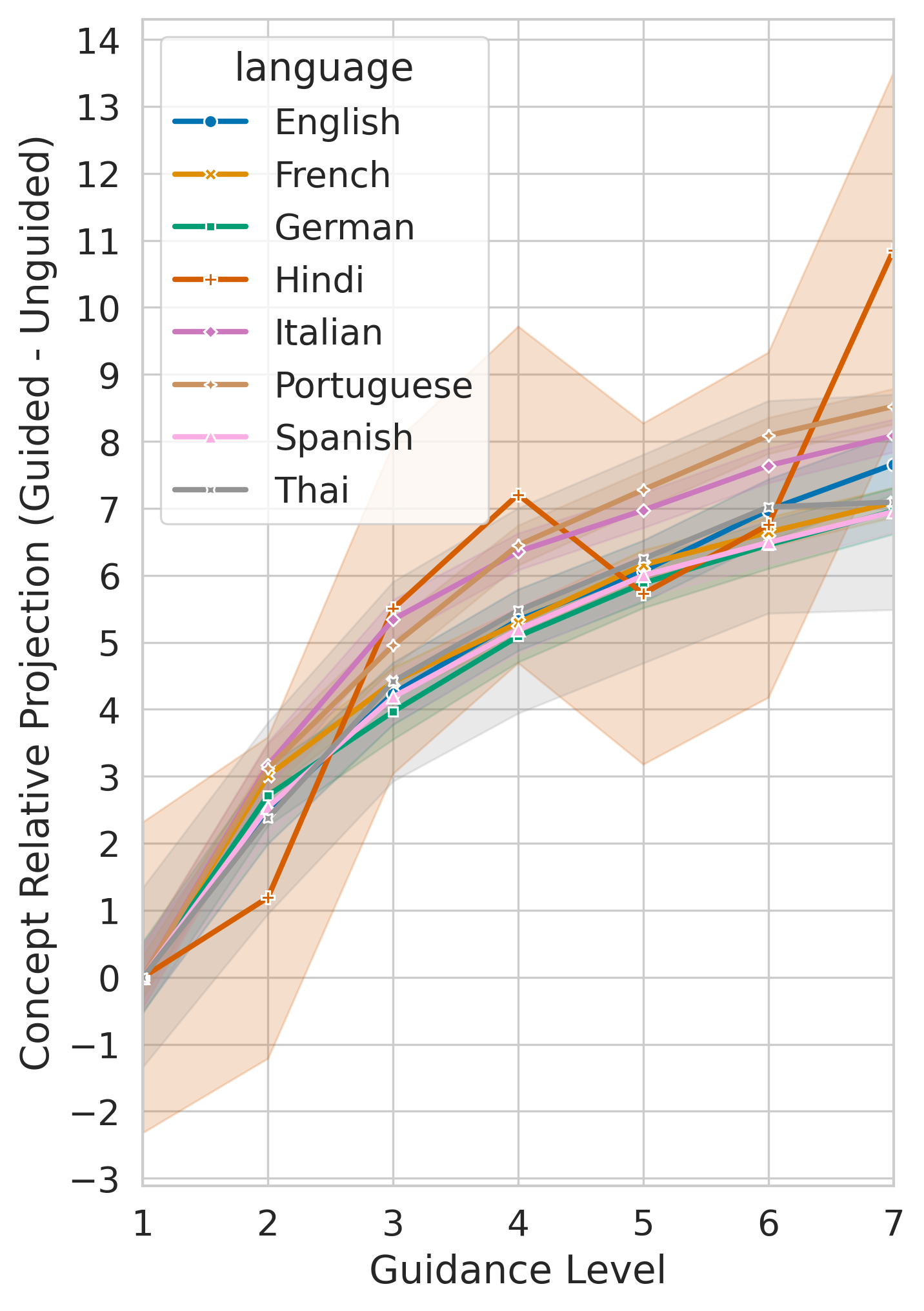}
    \caption{Growth of steering effect for the 8 languages supported by Llama 3.1 8B Instruct using top-$k$ guided generation. Rescaled for visibility.}
    \label{fig:guided_generation_language_topk_ablation}
\end{figure}

\begin{figure}[!ht]
    \centering
    \includegraphics[width=\linewidth]{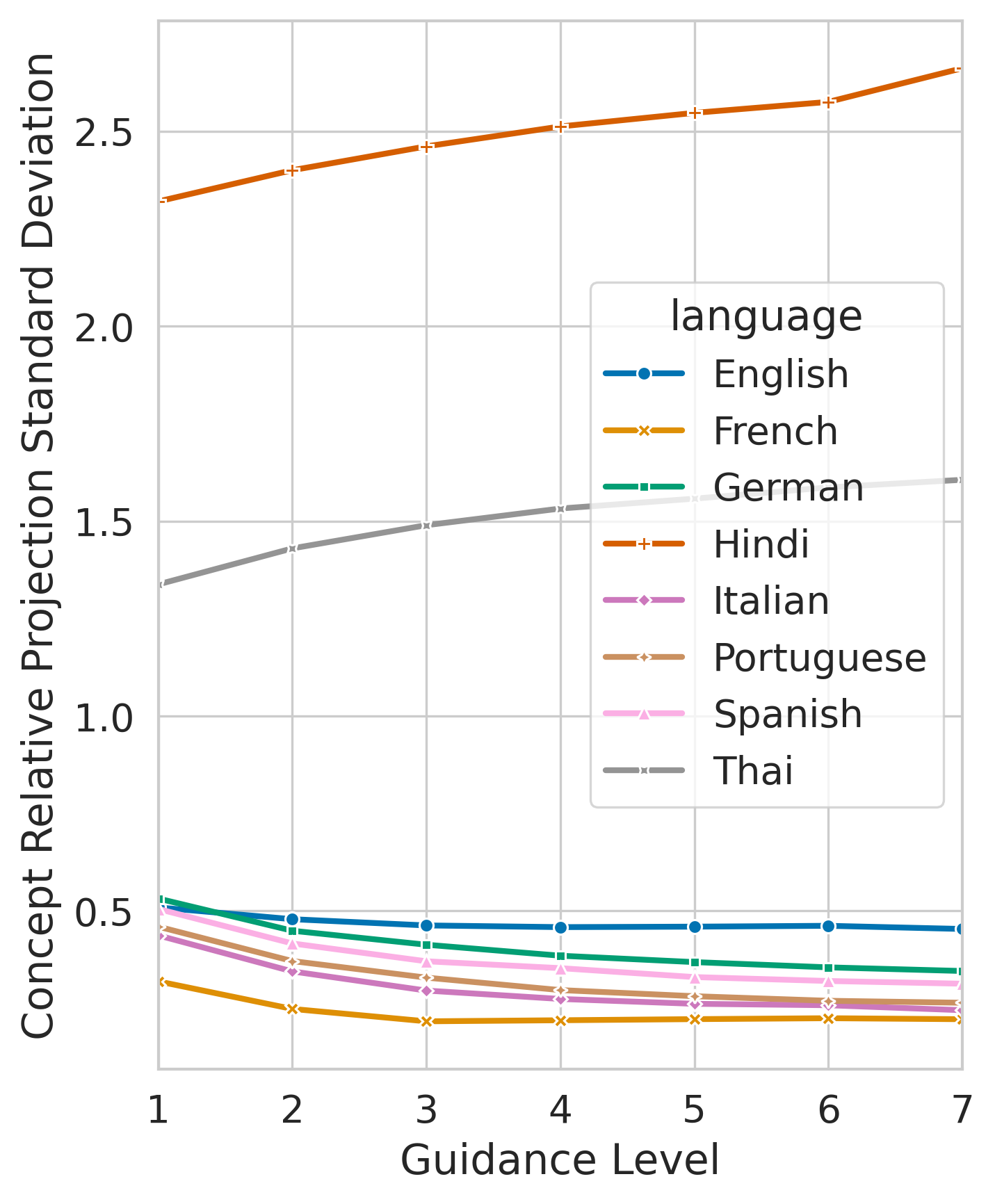}
    \caption{Growth of steering effect standard deviation for the 8 languages supported by Llama 3.1 8B Instruct using top-$k$ guided generation.}
    \label{fig:guided_generation_language_topk_ablation_std}
\end{figure}

We analyze the strength of the steering effect with respect to the $k$ factor. \Cref{fig:guided_generation_language_topk_ablation} shows that actually most languages are similarly affected on average, but as visible at \cref{fig:guided_generation_language_topk_ablation_std}, the standard deviation of the steering effect is higher for Hindi and Thai, which show a noisy pattern, possibly due limitations on the model's own capacity at handling these languages. 

\end{document}